\documentclass[sigplan,nonacm]{acmart}
\usepackage{geometry}
\usepackage{enumitem}
\usepackage{cleveref}
\usepackage{subcaption}
\usepackage{siunitx}
\usepackage{graphicx}

\AtBeginDocument{%
  }

\setcopyright{acmlicensed}
\copyrightyear{2024}
\acmYear{2024}
\acmDOI{10.1145/XXXXXXX.XXXXXXX}

\acmConference[ASPLOS '25]{31st IEEE International Symposium on High-Performance Computer Architecture}{March 2025}{Edinburgh, Scotland}

\acmISBN{978-1-4503-XXXX-X/2025/03}

\begin{document}

\title{AGFT: An Adaptive GPU Frequency Tuner for Real-Time LLM Inference Optimization}

\author{Zicong Ye}

\affiliation{%
  \institution{The Hong Kong University of\\
Science and Technology\\
(Guangzhou)
}
  \city{Guangdong}
  \state{Guangzhou}
  \country{China}
  \
}
\email{zye085@connect.hkust-gz.edu.cn}

\author{Kunming Zhang}

\affiliation{%
  \institution{The Hong Kong University of\\
Science and Technology\\
(Guangzhou)}
  \city{Guangdong}
  \state{Guangzhou}
  \country{China}
}
\email{kzhang519@connect.hkust-gz.
edu.cn}

\author{Guoming Tang*}
\affiliation{%
  \institution{The Hong Kong University of\\
Science and Technology\\
(Guangzhou)}
  \city{Guangdong}
  \state{Guangzhou}
  \country{China}
}
\email{a
guomingtang@hkust-gz.edu.cn}

\renewcommand{\shortauthors}{Chen et al.}

\begin{abstract}
The explosive growth of interactive Large Language Models (LLMs) has placed unprecedented demands for low latency on cloud GPUs, forcing them into high-power modes and causing escalating energy costs. Real-time inference workloads exhibit significant dynamic volatility, presenting substantial energy-saving opportunities. However, traditional static or rule-based power management strategies struggle to exploit these opportunities without compromising peak performance.

To address this challenge, we propose AGFT (An Adaptive GPU Frequency Tuner), a framework that employs online reinforcement learning to autonomously learn an optimal frequency tuning policy. By monitoring real-time features like request load and latency, AGFT utilizes fine-grained frequency control for precise adjustments and intelligent action space pruning for stable, efficient decision-making. This creates a robust, automated energy management solution.

We comprehensively evaluated AGFT in an environment simulating realistic, fluctuating inference requests. The experimental results demonstrate that AGFT successfully saves 44.3\% of GPU energy consumption while introducing a minimal performance latency overhead of under 10\%. This achievement translates into a comprehensive Energy-Delay Product (EDP) optimization of up to 40.3\%, clearly showing that our framework can significantly enhance the energy efficiency and economic benefits of existing LLM inference clusters without compromising service quality.
\\
\\
\end{abstract}



\received{15 November 2024}
\received[revised]{10 December 2024}
\received[accepted]{20 December 2024}

\maketitle

\section{Introduction}
\label{sec:introduction}

\textbf{Motivation.} The exponential growth of Large Language Model (LLM) inference workloads has presented unprecedented energy consumption challenges for modern data centers. For instance, a leading AI supercomputer such as xAI's Colossus is reported to have a power capacity of 280 megawatts (MW)  \cite{epoch:2024:trends}, with individual GPU like H100 continuously drawing 700 watts\cite{zhang:2023:h100whitepaper}. The demand for inference is particularly significant, potentially constituting over 90\% \cite{patel_characterizing_2024}of the total LLM compute cycles. As organizations deploy LLM services at scale, the energy costs and carbon footprint of these GPU-intensive inference operations have become critical concerns that demand innovative optimization strategies.

Although prior work has explored optimizing energy consumption through GPU frequency scaling, these methods often lack specificity for LLM inference scenarios\cite{zhang_improving_2024} and generally rely on predictive models based on offline training\cite{stojkovic_dynamollm_2024,kakolyris_throttllem_nodate}. Such methods suffer from two fundamental limitations. First, they are invasive and inflexible. Building an accurate offline model requires collecting massive amounts of production data covering diverse scenarios, which is not only a burdensome and costly process but also means that once the online workload patterns drift (e.g., due to changes in user behavior or model versions\cite{azure_public_datasets_nodate}), the pre-trained model quickly becomes obsolete and fails to adapt in real time\cite{wang2024survey,fedus2022switch}, leading to a significant degradation in optimization effectiveness. Second, these approaches raise data privacy concerns\cite{sharma_large_2023}. The process of uploading detailed operational traces, which may contain sensitive information, from production servers to a centralized platform for training poses severe privacy and security challenges in multi-tenant cloud environments. Therefore, a minimally intrusive, privacy-respecting, and continuously self-optimizing online solution is urgently needed.\\[3pt]
\textbf{Our work.} To address the aforementioned challenges that resolving the high energy consumption from real-time workload fluctuations in LLM inference services without compromising service quality, while avoiding the privacy risks and model obsolescence inevitable with traditional offline methods in dynamic environments—we must explore an online optimization path capable of real-time sensing and instantaneous adaptive decision-making. To this end, we conducted an extensive analysis of the real-time characteristics of Large Language Model (LLM) inference services, aiming to validate the strong correlation between the service's instantaneous state and optimal GPU configurations. Our findings show that the service’s real-time state—such as request queue depth, recent latency, and power consumption—is indeed strongly correlated with optimal GPU frequency configurations. This discovery reveals the immense potential for dynamic GPU frequency optimization through an agent capable of real-time sensing and immediate decision-making.

To transform this potential into a practical solution, we propose AGFT (An Adaptive GPU Frequency Tuner). The core advantage of this framework lies in its non-invasive, real-time online learning mechanism. AGFT models the dynamic control problem as an online reinforcement learning task, allowing it to learn on-the-fly in a live environment ("learning while running") and continuously adapt based on real-time feedback. This requires no offline data collection or pre-training, fundamentally avoiding the privacy risks of data migration and the problem of model obsolescence caused by environmental shifts. To ensure the precision and stability of real-time decision-making, AGFT integrates two core mechanisms: fine-grained frequency control and intelligent action space pruning, constituting a powerful and efficient automated real-time energy management solution.

In our evaluation, we replicated the real-time volatile workloads commonly found in production environments. The experimental results compellingly demonstrate that AGFT can reduce GPU energy consumption by a significant 44.3\% while introducing only minimal performance latency (a 9.3\% increase in TTFT and a 7.1\% increase in TPOT). More importantly, this optimization ultimately translates into a substantial 40.3\% improvement in the comprehensive Energy-Delay Product (EDP). These results clearly show that AGFT, as a fully online and real-time adaptive framework, can deliver significant and sustained energy efficiency improvements for LLM inference services without compromising service quality.\\\\

\textbf{Summary.} We make the following contributions:
\begin{itemize}[leftmargin=*,topsep=0pt, partopsep=0pt]
    \item We introduce the first frequency scaling system for LLM inference that achieves real-time optimization via a closed-loop, online learning framework, adapting to workload fluctuations while eliminating the need for offline modeling.
    
    \item We propose the first dynamic GPU frequency scaling method specifically designed for continuous batching in LLM inference.

    \item We propose the first minimally intrusive workload characterization method for LLM inference that fully preserves user privacy by operating without access to the user's prompt content or even its length.

    \item We characterize LLM inference workloads, identifying five prototypes each with a unique optimal frequency and a non-intrusive, 7-dimensional 'fingerprint' for real-time identification.

\end{itemize}

\section{Characteristics and Challenges of Modern Online LLM Inference Services
}
\label{sec:related}
The capabilities of modern online Large Language Model (LLM) inference services are advancing rapidly, yet their complex internal mechanisms pose unique challenges to optimizing the energy efficiency of underlying GPU systems. This section will first characterize the key features of these services, with a focus on their dynamic inference patterns and workloads, and then elucidate how these factors constitute the core bottlenecks for hardware optimization in LLM inference tasks.
\subsection{Shift in inference patterns } 
\vspace{2mm}
\textbf{Static batching.} The inference process of a Large Language Model (LLM) consists of two main phases: prefill and decode\cite{patel_splitwise_2024,kwon_efficient_2023}. In traditional static batching, a group of requests is processed together. The batch first enters a compute-bound prefill phase, where all input prompts are processed in parallel, making its performance highly sensitive to GPU core frequency\cite{pang_optimizing_2025,kamath_pod-attention_2025,cheng_kunserve_2025,agrawal_evaluating_2025}. Subsequently, the batch moves to a memory-bound decode phase, where the system generates one token for all ongoing requests in each iteration\cite{patel_splitwise_2024,kakolyris_throttllem_nodate,kwon_efficient_2023}. The entire process concludes only when the longest sequence in the batch is complete, at which point all results are returned simultaneously\cite{kwon_efficient_2023}. This clear distinction between computational characteristics is a key consideration for GPU optimization in this mode\cite{Posta2025DeepDive}.\\[3pt]
\textbf{Continuous batching. }Unlike traditional static batching, continuous batching is a more dynamic scheduling mechanism\cite{Yu2022Orca} that is key to enabling the streaming output of online LLM services.

The core of this mechanism is that the inference server maintains a continuous operational loop without waiting to form a complete batch\cite{kwon_efficient_2023}. As soon as a new request arrives, it can be immediately added to the current batch for prefilling \cite{NVIDIA_TRTLLM_Batching}, while other requests in the batch continue to be decoded in parallel. More importantly, when any request finishes its generation, its resources are instantly released and allocated to a new, waiting request. This "come-and-go" strategy \cite{Yu2022Orca} eliminates the GPU idle time caused by waiting for the longest sequence\cite{Anyscale2023ContinuousBatching}, thereby significantly improving the system's overall throughput and resource utilization.\\[3pt]\begin{figure}
    \centering
    \includegraphics[width=1\linewidth]{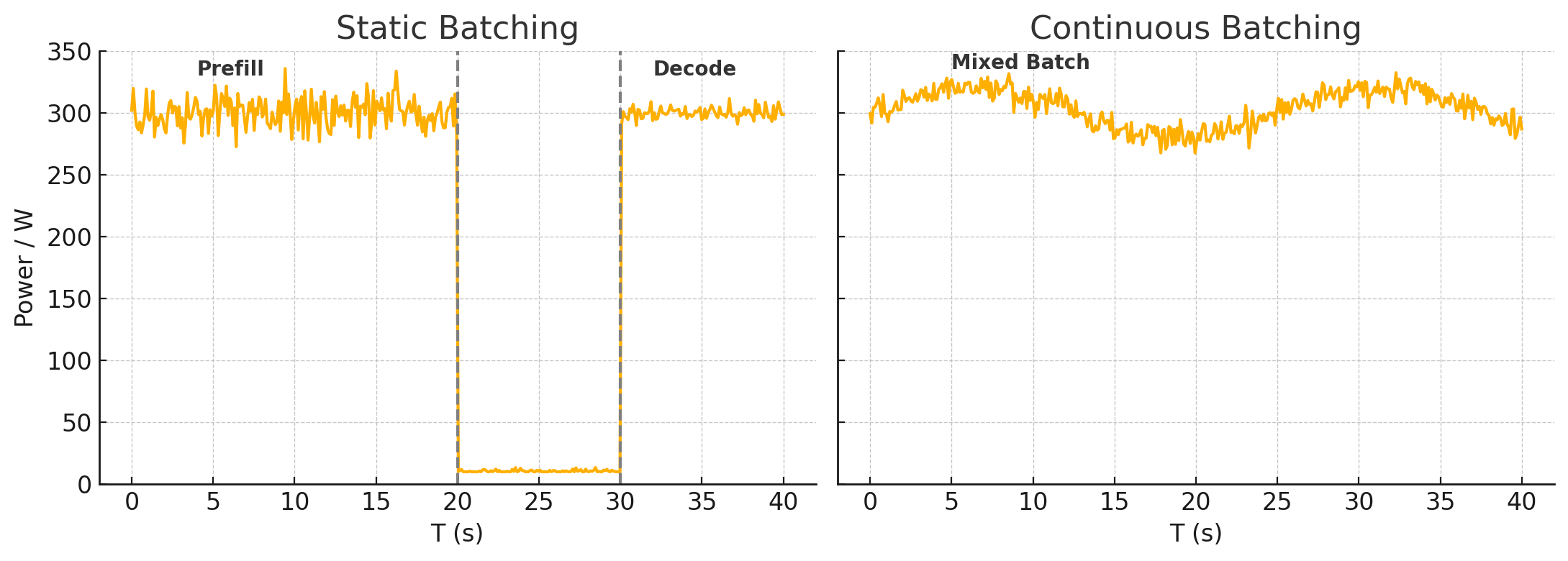}
    \caption{Power variation of an LLM during inference tasks with static and continuous batching.}
    \label{Figure:1}
\end{figure}
\textbf{Challenge. }The evolution of batching technology has significantly improved the inference throughput of LLMs. However, the mixture of the compute-intensive prefill phase and the memory-intensive decode phase presents a unique challenge for Dynamic Voltage and Frequency Scaling (DVFS) technology.

To investigate the impact of different batching modes on hardware resource utilization, we conducted experiments on an NVIDIA A800 single-card server with the open-source Llama2-7B model under an equivalent request rate. Figure 1 illustrates the real-time GPU power changes during inference using both static and continuous batching modes.

In Static Batching mode (\Cref{Figure:1}, left), the compute-intensive prefill (approx. 280-325W) and stable decode phases (approx. 300W) have distinct power signatures, allowing for easy phase identification via power monitoring. In contrast, Continuous Batching mode (\Cref{Figure:1}, right) interleaves these operations, resulting in a constantly fluctuating high-power state (approx. 270-325W) that masks individual phase characteristics. This "averaging" effect makes it extremely difficult to identify the computational phase (i.e., prefill or decode) from real-time hardware telemetry, posing a core challenge for formulating an optimal DVFS strategy.\\
\subsection{User Privacy and Reduced System Intrusiveness for Online LLM Inference Services}
\textbf{User Data Encryption. }With the growing awareness of user privacy and security, protecting "data-in-use" has become a critical challenge for online LLM inference services. While traditional encryption protects data at-rest and in-transit, data must be processed in plaintext\cite{Zhu2024Confidential} in memory during computation, creating a vulnerability where service providers could potentially access sensitive user information.

\begin{figure}
    \centering
    \includegraphics[width=1\linewidth]{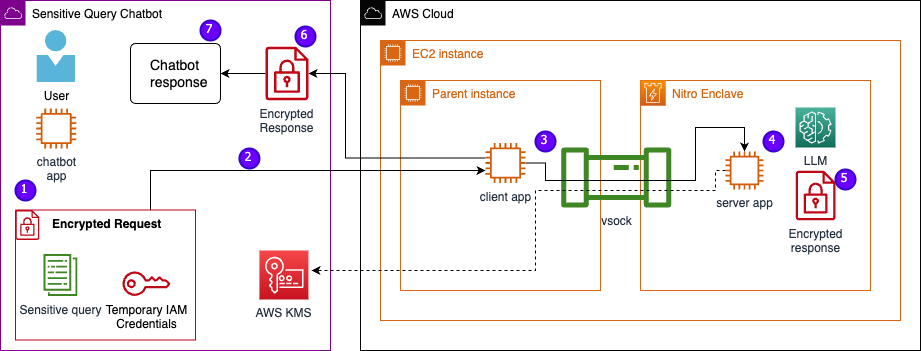}
    \caption{AWS Nitro Enclaves Overview}
    \label{fig:2}
\end{figure}

To address this leakage risk, confidential computing solutions using Trusted Execution Environments (TEEs) have emerged. As illustrated in \Cref{fig:2}, architectures like AWS Nitro Enclaves\cite{AWS_Nitro_Enclaves_Docs} are designed to prevent even the service provider from accessing plaintext data. In this operational model\cite{AWS_Blog_LLM_Nitro}, a user's encrypted query is sent to an isolated Enclave where all sensitive operations—such as decryption, inference, and re-encryption—are completed within a secure "black box." This establishes a non-intrusive boundary, altering the provider's role to a mere facilitator of computation\cite{Priebe2018Survey,Brasser2019Survey} who has no access to plaintext data, thereby guaranteeing the confidentiality of user requests and model responses\cite{AWS_Nitro_Enclaves_Docs}.\\[3pt]
\textbf{Challenge.} The core challenge for advanced DVFS systems is to reconcile energy optimization with the constraints of a non-intrusive privacy model. Effective energy optimization hinges on accurately predicting the workload in real-time. As research indicates, a key factor is the dynamic growth of the KV cache, which correlates with the number of generated tokens\cite{kakolyris_throttllem_nodate}. Forecasting this number would enable proactive frequency planning, but the very nature of the non-intrusive architecture creates a fundamental conflict.

To guarantee privacy, the service provider is intentionally blinded from the user’s prompt content. This information isolation makes it impossible to semantically predict the generation task's length or complexity, depriving the DVFS system of the crucial forward-looking data needed for its decision-making. Consequently, achieving an optimal energy-performance trade-off becomes extremely difficult, establishing a core research problem in designing systems that are both high-performance and non-intrusive.
\subsection{Energy-Delay Product and Optimization}
\textbf{The Energy-Delay Product (EDP).} Given the challenges of mixed computational phases and real-time workload volatility, evaluating the efficiency of LLM inference services requires a metric that transcends a singular focus on either performance or energy. Optimizing for latency alone leads to excessive power consumption, while optimizing solely for energy results in unacceptable performance degradation\cite{Maliakel2025Investigating,Gao2024Investigating}. The Energy-Delay Product (EDP), defined as EDP = Energy × Delay, serves as a crucial, balanced metric\cite{Shao2023DOSA,Kwon2023HighLight}. It holistically captures the trade-off between performance and power, with a lower EDP value indicating a more desirable state of overall system efficiency.\\[3pt]
\textbf{Challenge.} The central optimization challenge for LLM inference services, therefore, becomes how to dynamically minimize the EDP in real-time. This is a formidable task due to the confluence of issues previously discussed. The unpredictable mixture of compute-bound and memory-bound operations in continuous batching obscures the relationship between frequency and system efficiency\cite{patel_characterizing_2024,kamath_pod-attention_2025}. Furthermore, the non-stationary nature of real-world workloads means that any statically determined optimal operating point, or "sweet spot" for the EDP, is fleeting\cite{stojkovic_dynamollm_2024}. Compounded by privacy constraints that prevent predictive optimization based on request content\cite{MNN-AECS2025}, finding the ideal GPU frequency to consistently achieve the lowest possible EDP under these fluctuating and opaque conditions remains a critical research problem\cite{Qiu2024muServe,kakolyris_throttllem_nodate}.
\subsection{Real-world workload dynamics}
\begin{figure}
    \centering
    \includegraphics[width=1\linewidth]{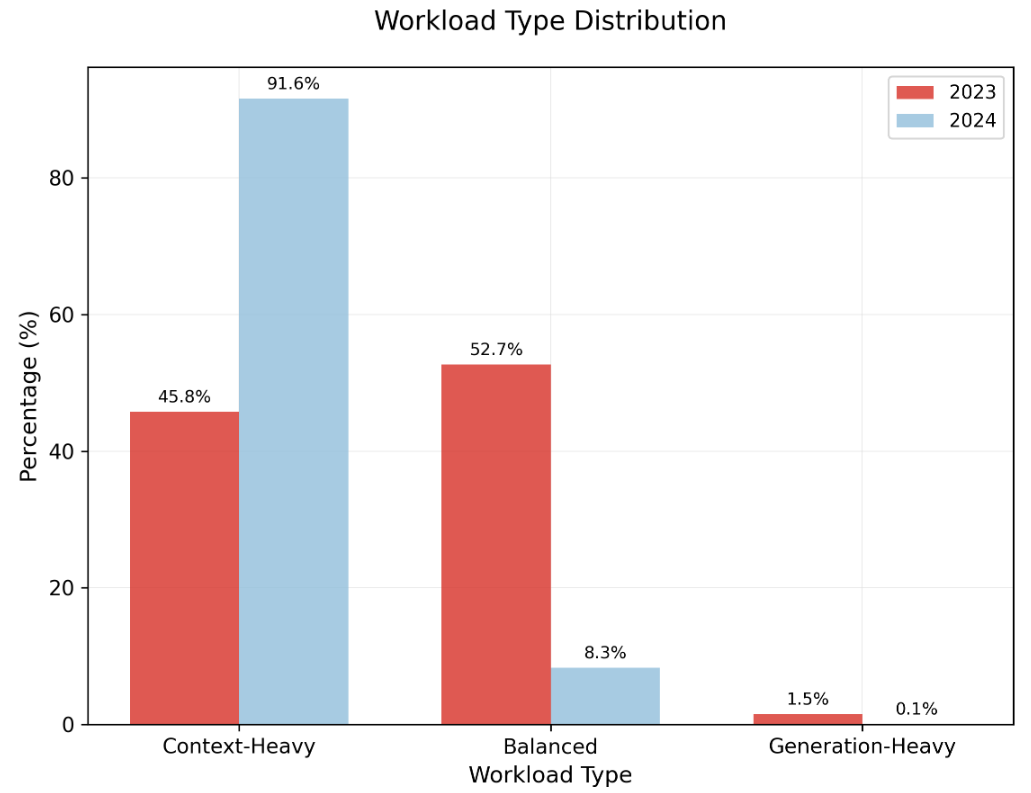}
    \caption{Yearly evolution of workload types (2023 vs. 2024).}
    \label{fig:3}
\end{figure}

\begin{figure}
    \centering
    \includegraphics[width=1\linewidth]{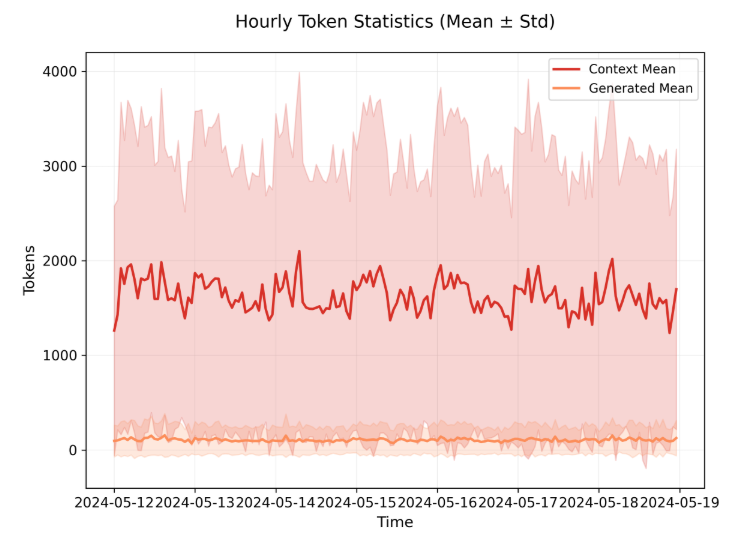}
    \caption{Short-term workload dynamics over a one-week period (hourly mean ± std).}
    \label{fig:4}
\end{figure}

To analyze the evolving load characteristics of production LLM services, we leverage two official call trace datasets from a large-scale inference service on Microsoft Azure, spanning usage in 2023 and 2024\cite{AzureLLMDataset}. These datasets, analyzed in \cite{wang2024burstgpt}, precisely record the timestamp, input token count (context), and output token count (generated) for each inference task.\\[3pt]
\textbf{Yearly Evolution. }The long-term evolutionary trend of the workload is highly significant. \Cref{fig:3} illustrates the drastic change in the workload type distribution over one year. In 2023, the workload was primarily composed of Balanced and Context-Heavy types (52.7\% and 45.8\%). By 2024, a paradigm shift occurred: Context-Heavy workloads surged to become dominant, constituting 91.6\% of the total. Consequently, Balanced workloads dropped to 8.3\%, and Generation-Heavy requests, already rare (1.5\%), nearly vanished (0.1\%)\cite{wang2024burstgpt, AzureLLMDataset}.\\[3pt]
\textbf{Weekly Variations. }Beyond long-term evolution, the workload within Azure's production environment exhibits intense short-term uncertainty. \Cref{fig:4} depicts hourly average token statistics over one week in May 2024. The average input (Context) tokens show extreme volatility, frequently oscillating between 1,200 and 2,100, with a standard deviation upper bound often exceeding 3,500. In stark contrast, the average output (Generated) tokens remain consistently low and stable at approximately 100 to 200\cite{wang2024burstgpt, AzureLLMDataset}.\\[3pt]
\textbf{Challenge. }This analysis of Azure's production traces reveals a key characteristic of real-world LLM workloads: a high degree of non-stationarity. This manifests not only in fundamental structural shifts on a yearly scale but also in the unpredictable, dynamic variations of input sizes on an hourly scale.
This dynamic nature of the workload poses a severe challenge to existing LLM inference optimization methods, especially frequency scaling strategies that rely on offline profiling. For instance, prior works (such as DynamoLLM) typically build GPU performance and power models by running tests under a set of predefined, static workload conditions\cite{stojkovic_dynamollm_2024}. The core assumption of such an approach is that the relationship between workload characteristics and hardware performance remains stable over time. Our analysis, however, demonstrates that this assumption does not hold in real-world production environments. As online load characteristics continuously evolve, any offline model built on historical data will inevitably become "stale" quickly, leading the system to make suboptimal frequency decisions that ultimately either sacrifice performance or waste energy. Therefore, to achieve robust energy-efficiency optimization amidst these constantly changing loads, there is an urgent need for a new paradigm capable of online learning of workload features and adaptively adjusting strategies in real-time\cite{kwon2023vllm,agrawal2024sarathi,sun2024llumnix}.

\begin{figure}[t!]
  \centering 

  \begin{subfigure}{1\linewidth}
    \centering
    \includegraphics[width=0.8\linewidth, keepaspectratio]{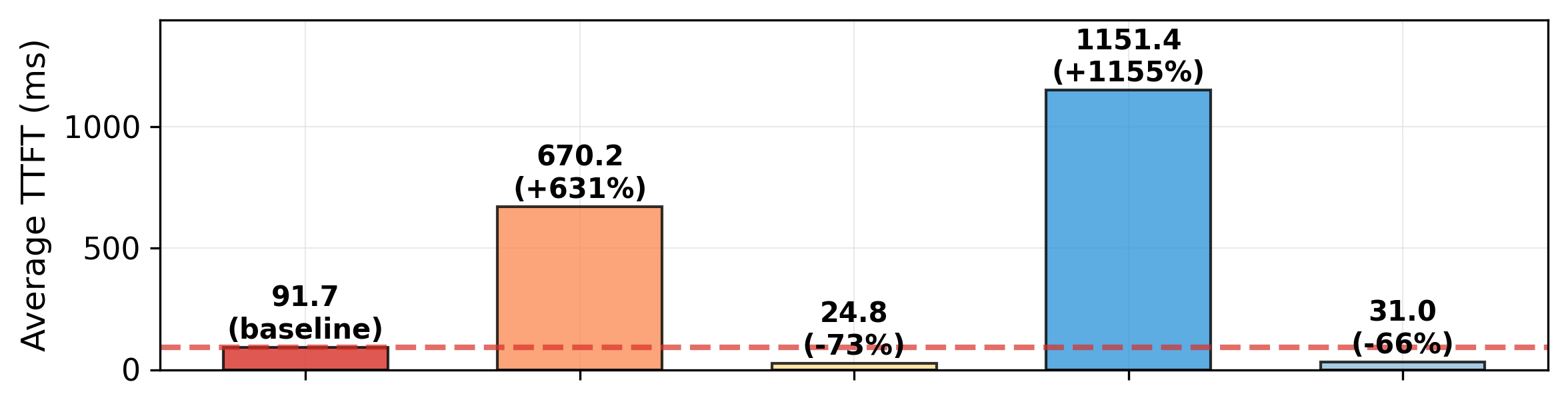}
    \caption{Average Time to First Token}
  \end{subfigure}
  
  \vspace{1em} 

  \begin{subfigure}{1\linewidth}
    \centering
    \includegraphics[width=0.8\linewidth, keepaspectratio]{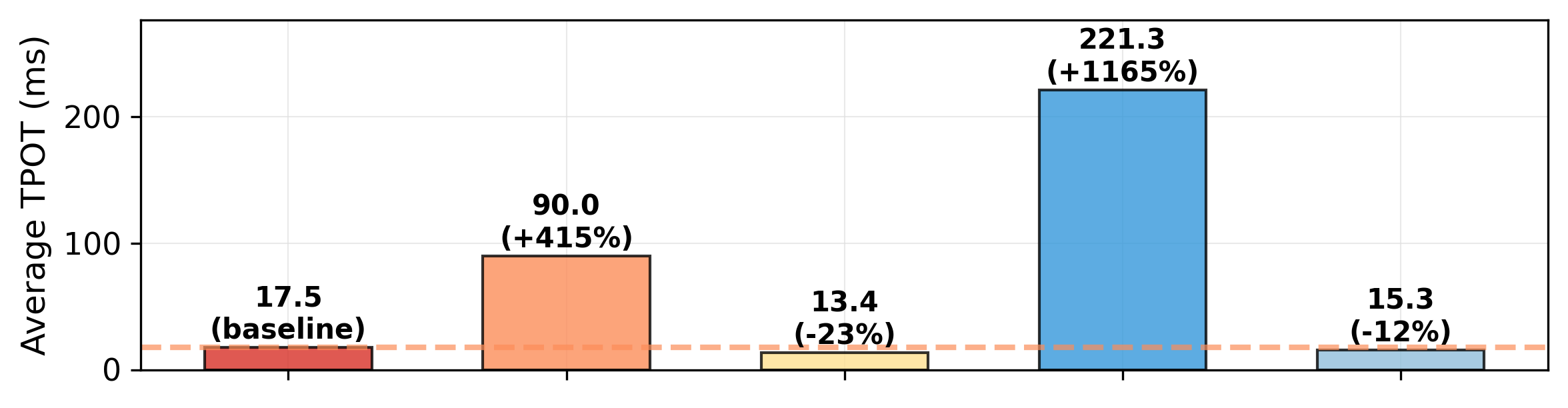}
    \caption{Average Time per Output Token}
  \end{subfigure}
  
  \vspace{1em}

  \begin{subfigure}{1\linewidth}
    \centering
    \includegraphics[width=0.8\linewidth, keepaspectratio]{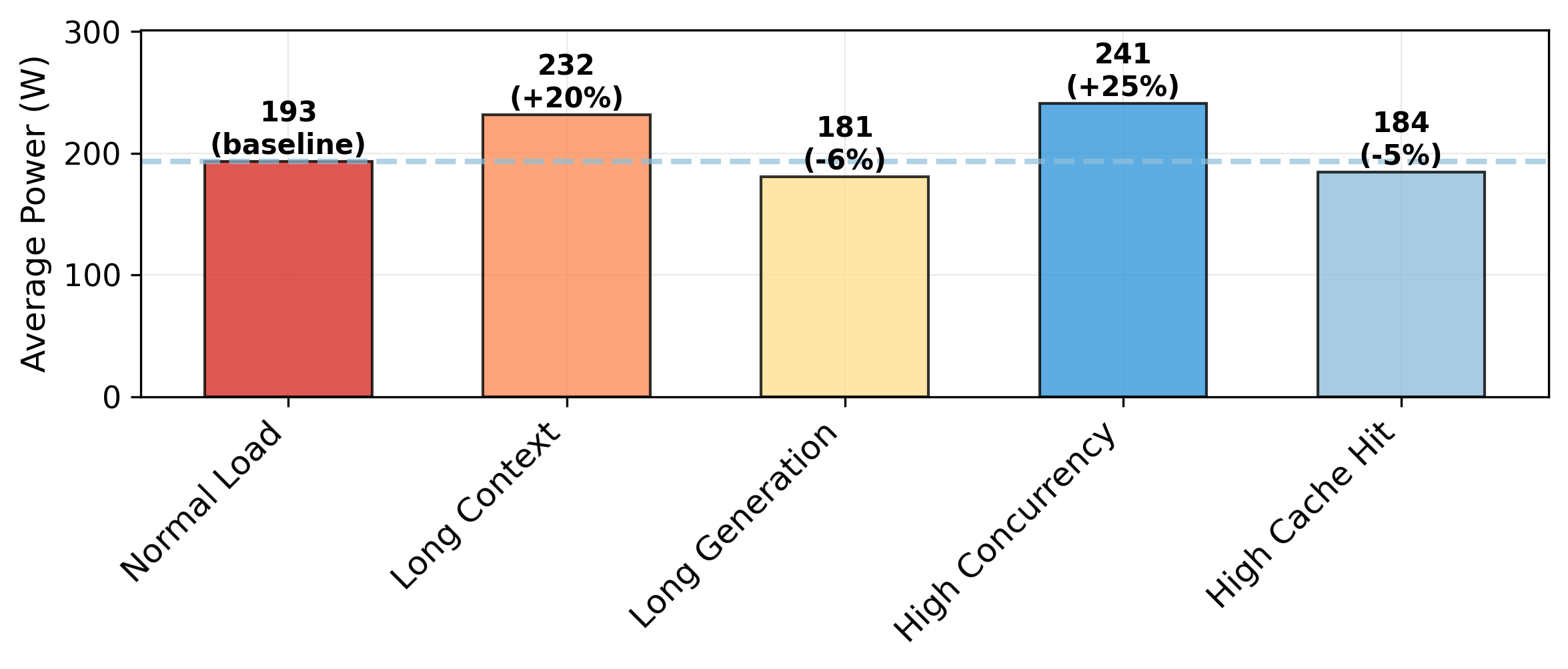}
    \caption{Average Power Consumption}
    \label{fig:5c}
  \end{subfigure}

  \caption{Performance and Power Profiling across Different Workload Prototypes}
  \label{fig:performance_power_analysis}
\end{figure}

  \begin{table*}[t]
  \centering
  \begin{tabular}{lcccc}
  \toprule
  \textbf{Workload} & \textbf{Context} & \textbf{Generation} & \textbf{Concurrency} & \textbf{Prompt Templates} \\
  \midrule
  Normal Load & 256-1024 & 100-350 & 1x & 500 \\
  Long Context & 1024-8192 & 1-100 & 1x & 500 \\
  Long Generation & 1-256 & 350 & 1x & 500 \\
  High Concurrency & 256-1024 & 100-350 & 5x & 500 \\
  High Cache Hit & 256-1024 & 100-350 & 1x & 5 \\
  \bottomrule
  \end{tabular}
  \caption{Experimental Configuration for Different Workloads}
  \label{tab:workload_config}
  \end{table*}

\section{Privacy-Preserving Real-time Workload Characterization and Identification}
\label{sec:background}
The challenges established in the preceding chapter—opaque operational states from continuous batching, stringent privacy constraints, and profound workload non-stationarity which render traditional static optimizations ineffective. This necessitates a paradigm shift towards the dynamic, real-time characterization of workload properties using only non-intrusive signals. Achieving such precise and energy-saving frequency control first requires a profound understanding of the underlying workload, which is the central focus of this chapter \cite{Yu2022Orca,kwon2023vllm}. Accordingly, we will establish a framework for this dynamic characterization by sequentially demonstrating: (1) the significant performance and power disparities among different workload prototypes\cite{fernandez2025energy,vellaisamy2025characterizing}; (2) how these disparities result in a unique optimal GPU frequency for each prototype\cite{gairola2025dvfsgpt,kakolyris_throttllem_nodate}; and (3) a non-intrusive method to efficiently identify these prototypes in real-time using a 7-dimensional feature vector\cite{gao2024workloadgpt,ivanov2024realtime}.

To systematically evaluate the system's behavior under diverse operational conditions, we constructed five distinct workload prototypes. These prototypes are defined by manipulating four key parameters, each chosen to probe a fundamental aspect of the LLM inference process.

First, foundational work has established that the input context length and output generation length are the primary determinants of a workload's composition, governing the balance between the compute-intensive prefill and memory-bound decode phases \cite{stojkovic_dynamollm_2024}. We therefore select these as our initial parameters for variation. Second, the concurrency level is adjusted to simulate varying request pressures on the server, a critical factor for assessing system-level throughput and resource contention under realistic load scenarios. Third, to account for the impact of modern architectural optimizations, we manipulate the prompt template pool size. This directly models the effectiveness of prefix caching mechanisms, such as the one employed by vLLM. A small, repetitive template pool (e.g., 5 templates) simulates a high KV cache hit rate by maximizing prefix reuse, thus significantly reducing prefill overhead, whereas a large pool represents workloads with low cache locality \cite{kwon2023vllm, yuzuguler2025preserve}.\Cref{tab:workload_config} details the specific parameter configurations for each workload prototype established for our experiments.

\subsection{Performance and Power Differentiation of Workload Prototypes}

\begin{figure*}
    \centering
    \includegraphics[width=1\linewidth]{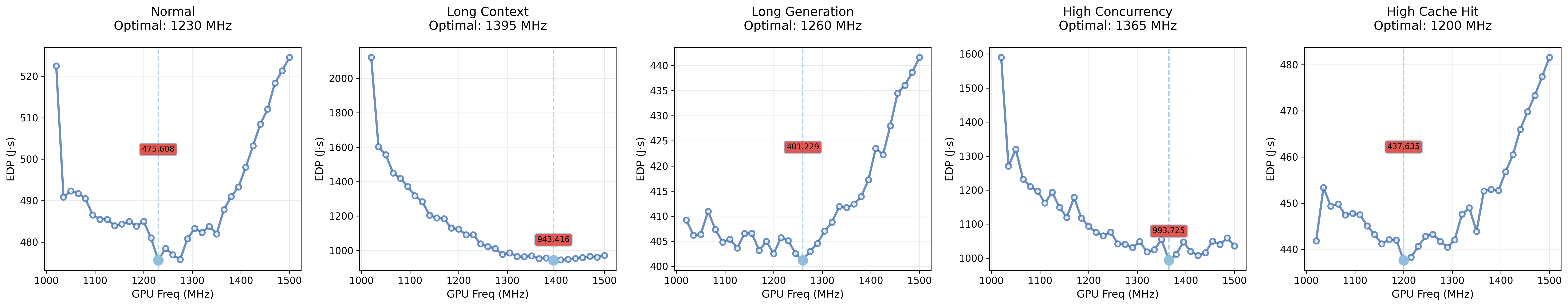}
    \caption{Optimal GPU frequency varies across different workload prototypes. The plots show the Energy-Delay Product (EDP) versus GPU frequency, with the minimum EDP point highlighted.}
    \label{fig:6}
\end{figure*}

\textbf{Effect On Performance. }\Cref{fig:performance_power_analysis} presents an evaluation of the system's performance and power consumption across five distinct workloads, where each evaluation consisted of a 5000-task round. With the "Normal Load" scenario serving as the performance baseline, our results indicate that "Long Context" and "High Concurrency" conditions lead to significant performance degradation. Notably, the "High Concurrency" workload caused the Average Time to First Token (TTFT) and Average Time per Output Token (TPOT) to surge by 1153\% and 116\%, respectively. In contrast, both the "Long Generation" and "High Cache Hit" workloads substantially enhanced performance, with the former achieving a 73\% reduction in TTFT.\\[3pt]
\textbf{Effect On Power. }The system's power consumption profile is highly correlated with its performance latency trends (TTFT/TPOT). As shown in \Cref{fig:5c}, under the "High Concurrency" and "Long Context" workloads, which cause severe performance degradation, the system's average power consumption also surges. Notably, the "High Concurrency" scenario not only exhibits the highest latency but also reaches a peak power consumption of 241 W, a 25\% increase compared to the 193 W baseline. This clearly indicates that intensive context switching and resource contention are primary sources of the system's energy consumption. On the other hand, the low-latency scenarios, namely "Long Generation" and "High Cache Hit," demonstrate superior energy efficiency. Their power consumption levels are both below the baseline (181 W and 184 W, respectively). This is primarily attributed to effectively avoiding or reducing the most computationally expensive initial prefill stage. "Long Generation" shifts the main workload to the more power-stable token-by-token decoding process, while "High Cache Hit" leverages a caching mechanism to bypass most of the prefill computation. Both approaches successfully avoid the power peaks associated with processing large-scale parallel computation tasks.\\[3pt]
\textbf{Hypothesis.} Based on the preceding quantitative analysis of system performance and power consumption, we propose a core hypothesis: there is no single "globally optimal" GPU core frequency applicable to all LLM inference scenarios. Instead, the optimal performance-power trade-off is closely tied to the characteristics of the workload. This implies that an intelligent inference system should adopt a dynamic, workload-aware frequency scaling strategy to achieve different optimization objectives.Specifically, our hypothesis can be broken down into the following two points:
\begin{enumerate}[label=\arabic*),topsep=2pt]
    \item \textbf{Computation-Intensive Workloads Demand Higher Frequencies:} For workloads highly sensitive to latency, such as ``High Concurrency'' and ``Long Context,'' the optimal frequencies are pushed to a higher range of 1365--1395~MHz. In these scenarios, performance is the primary bottleneck; a lower frequency would cause a sharp increase in latency that dominates the Energy-Delay Product (EDP) degradation. Thus, a higher frequency is required to ensure performance and achieve the best energy-delay balance.

    \item \textbf{Efficiency-Oriented Workloads Favor Lower Frequencies:} Conversely, for less computationally demanding workloads like ``High Cache Hit,'' ``Normal Load,'' and ``Long Generation,'' the optimal frequencies are found in a lower 1200--1260~MHz range. For these workloads, where performance is less constrained (e.g., due to effective caching or being decode-bound), the EDP is more sensitive to energy savings. A moderate frequency reduction thus yields a better overall efficiency by significantly cutting power consumption with only a minor impact on latency.
\end{enumerate}

\subsection{Optimal Frequencies for Different Workloads. }
To find the energy-efficiency ``sweet spot'' for different LLM inference scenarios, we conducted a comprehensive GPU frequency sweep across five typical workloads using the Energy-Delay Product (EDP) as the primary metric. In the experiment, we iterated through all core frequencies of an NVIDIA A6000 GPU from 210~MHz to 1800~MHz, with a step size of 15~MHz. At each frequency point, we measured the total energy consumption and total time delay required to complete 5000 inference requests and calculated the corresponding EDP value ($EDP = \text{Energy} \times \text{Delay}$, where lower is better).\\[3pt]
\textbf{Characterization. }The experimental results are presented in \Cref{fig:6}. The figure clearly reveals that the EDP for each workload exhibits a frequency-sensitive ``U-shaped'' curve, which implies that the optimal energy-efficiency balance point exists at a specific frequency, rather than at the highest or lowest available frequency. Our core findings are as follows:
\begin{enumerate}[label=\arabic*),topsep=2pt]
    \item \textbf{Computation-Intensive Workloads Demand Higher Frequencies:} For workloads highly sensitive to latency, such as ``High Concurrency'' and ``Long Context,'' the optimal frequencies are pushed to a higher range of 1365--1395~MHz. In these scenarios, performance is the primary bottleneck; a lower frequency would cause a sharp increase in latency that dominates the Energy-Delay Product (EDP) degradation. Thus, a higher frequency is required to ensure performance and achieve the best energy-delay balance.

    \item \textbf{Efficiency-Oriented Workloads Favor Lower Frequencies:} Conversely, for less computationally demanding workloads like ``High Cache Hit,'' ``Normal Load,'' and ``Long Generation,'' the optimal frequencies are found in a lower 1200--1260~MHz range. For these workloads, where performance is less constrained (e.g., due to effective caching or being decode-bound), the EDP is more sensitive to energy savings. A moderate frequency reduction thus yields a better overall efficiency by significantly cutting power consumption with only a minor impact on latency.
\end{enumerate}
\textbf{Research Question.} To enable fine-grained dynamic GPU frequency optimization, the system must first accurately identify its current workload mode. Although the most direct approach is to analyze the user's prompt request itself—for instance, by checking its length to determine a "Long Context" mode, or by parsing its content to estimate the reply length and thus identify a "Long Generation" load—this method is highly invasive and raises serious privacy and security concerns. In particular, directly reading and parsing prompt content crosses an inviolable privacy red line and is unacceptable in any responsible application. Therefore, a core challenge arises: Can we find an indirect yet reliable method to effectively distinguish between different workloads, without reading user prompt content or relying on the analysis of individual request lengths?
\begin{figure}
    \centering
    \includegraphics[width=1\linewidth]{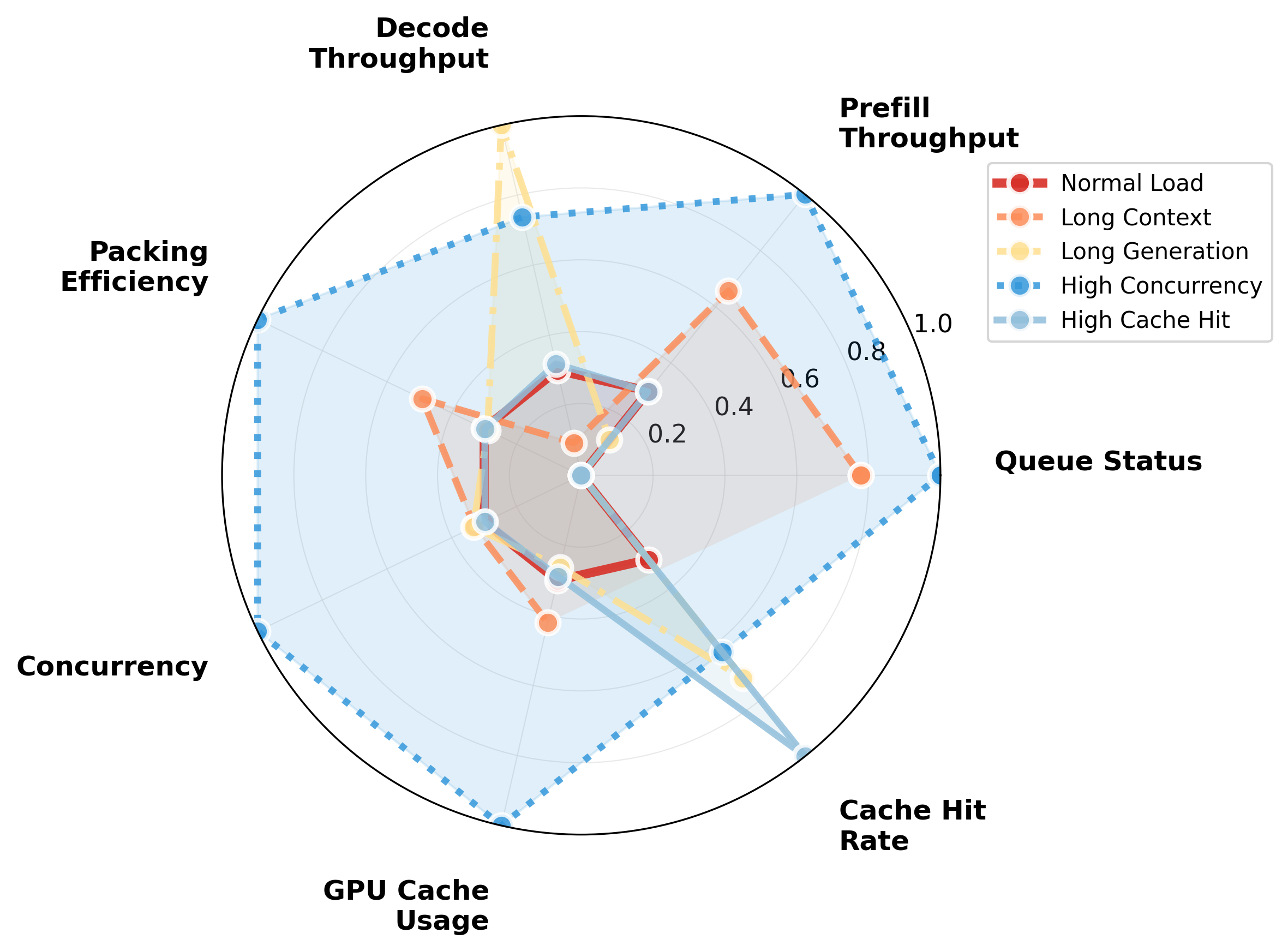}
    \caption{Radar chart visualization of the 7-dimensional feature fingerprints for the five typical workload prototypes. Each colored polygon represents the normalized average feature values for a specific workload.}
    \label{fig:7}
\end{figure}
\subsection{Workload Fingerprint}
To achieve effective workload identification while strictly adhering to the design principles of user privacy protection and minimal invasiveness, we propose a 7-dimensional feature vector based on macro performance indicators. The design of this feature set cleverly utilizes the intrinsic characteristics of the ``Continuous-Batching'' mechanism found in modern inference serving frameworks.

Unlike traditional methods that wait to form a complete batch, continuous batching allows the system to dynamically and continuously accept and process new requests. This results in a complex combination of multiple requests at different computation stages (such as prompt prefill and subsequent token decoding) constantly running on the GPU. This dynamically changing ``runtime state'' is exposed through a series of macro performance indicators that do not involve the specific content of any individual request, which is key to our minimally-invasive monitoring approach.\\[3pt]
\textbf{7-dimensional features. }In our research, we use the current mainstream vLLM framework as an example to specifically demonstrate how to capture these macro indicators and construct our 7-dimensional feature ``fingerprint''. The specific dimensions are as follows:

\begin{enumerate}
    \item \textbf{Queue Status (Has Queue):} A binary indicator to determine if there are currently requests waiting to be processed. It is the most direct signal of whether the system is under load.

    \item \textbf{Prefill Throughput:} The total number of prompt tokens processed by the system within a unit sampling period. This feature reflects the system's capacity for processing new inputs and is highly correlated with ``Long Context'' type workloads.

    \item \textbf{Decode Throughput:} The total number of tokens generated by the system within a unit sampling period. This feature primarily measures the model's content generation speed and is key to identifying ``Long Generation'' workloads.

    \item \textbf{Batching Efficiency (Packing Efficiency):} The average number of tokens processed per iteration within a batch. This metric measures the efficiency of the inference framework's dynamic batching mechanism and reflects how tightly requests are combined.

    \item \textbf{Concurrency:} The number of requests currently being processed by the system. This is a core load indicator that directly corresponds to ``High Concurrency'' scenarios.

    \item \textbf{GPU Cache Usage:} The percentage of GPU memory occupied by the key-value cache (KV Cache). This metric indicates the memory pressure exerted by the currently running tasks.

    \item \textbf{Cache Hit Rate:} The hit percentage of the system's prefix cache. This metric is the most effective feature for identifying ``High Cache Hit'' scenarios and is an aggregate statistic that does not expose any individual user's information.
\end{enumerate}
\begin{figure*}[t] 
    \centering 
    \includegraphics[width=0.9\textwidth] 
    {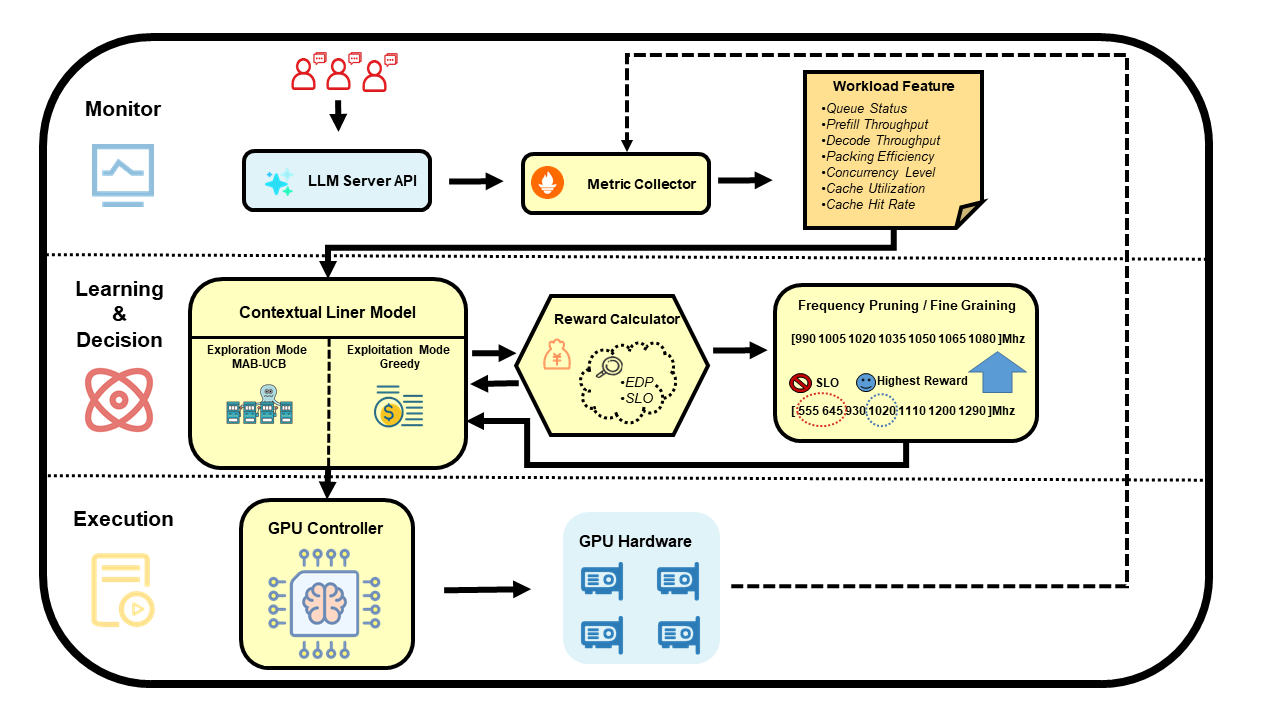} 
    \caption{The overall architecture of AGFT, detailing its core modules and their interactions with the vLLM service and GPU hardware. The system adaptively selects GPU frequencies to optimize the Energy-Delay Product (EDP).} 
    \label{fig:vAFT_architecture} 
\end{figure*}
\textbf{Proof. }To verify whether our proposed 7-dimensional feature vector can effectively serve as a "fingerprint" \cite{laor2022drawnapart, zhang2025fingerprinting, zhang2025gpu} to distinguish between different workloads, we conducted a feature visualization experiment. Consistent with the previous setup, we ran five typical workloads on an NVIDIA A6000 GPU in its default dynamic mode with an unlocked core frequency. Each workload consisted of 5000 inference requests. During this process, we continuously collected the instantaneous values of the 7-dimensional features within a 0.8-second sampling window and calculated the average value for each feature after each round of tasks. The use of low-level GPU performance counters for such detailed workload characterization is a common practice in modern GPU management systems \cite{wang2022gpoeo, aws2023monitoring, google2023monitoring}. To facilitate comparison on the same scale, we normalized these average values.

The experimental results are presented in \Cref{fig:7}. We plotted the normalized average values of the 7-dimensional features for each workload on a Radar Chart, thereby forming their unique "Workload Fingerprints." The figure clearly shows that the five workloads have distinct fingerprint shapes with high distinguishability, which strongly demonstrates the effectiveness of our selected feature set. Serving as the baseline for our analysis, the "Normal Load" (red solid line) exhibits the most balanced fingerprint shape across all dimensions and is positioned closer to the center, providing a frame of reference for the behavior of other specialized workloads. In contrast, the other workloads present distinct "biased" fingerprints: the "High Concurrency" load (blue dotted line) shows extremely sharp peaks on the "Concurrency" and "Queue Status" dimensions, which is characteristic of systems under high load with long request queues\cite{bento2024benchmarking, google2024autoscaling, qiu2024qlm}; the "Long Context" load (orange dashed line) is prominent on "Prefill Throughput" and "GPU Cache Usage", consistent with the compute-bound nature of the prefill phase and the large memory footprint of the KV cache in long-context scenarios\cite{agrawal2024sarathi, alibabacloud2024kv, dell2024memory}; the "High Cache Hit" load (light blue solid line) has a near-saturated value on the "Cache Hit Rate" dimension\cite{kwon_efficient_2023,nvidia2024kv,vllm2024prefix}; and the "Long Generation" load (yellow dash-dot line) mainly displays its characteristics on "Decode Throughput"\cite{kamath_pod-attention_2025,agrawal2024sarathi,agrawal2023sarathi}.

These unique and quantifiable fingerprints lay a solid foundation for building the machine learning-based workload identifier and dynamic decision-making system \cite{agrawal2024workload,wang2022gpoeo}, in subsequent chapters.

\section{AGFT Design}
\label{sec:methodology}
\textbf{Architecture.} Based on our insights, we propose AGFT, an adaptive GPU frequency optimization framework. This framework is specifically designed for Large Language Model (LLM) inference services that adopt Continuous-Batching technology\cite{kwon2023vllm,Yu2022Orca}. AGFT aims to solve the unique challenge of balancing performance and energy consumption under the dynamic and variable nature of inference workloads. While strictly adhering to Service Level Objectives (SLOs), AGFT significantly reduces the GPU’s overall energy consumption by optimizing the Energy-Delay Product (EDP) in real-time, thereby achieving a better energy-efficiency ratio without impacting service quality. The entire system operates on a single inference node, interacting directly with the GPU hardware and the underlying inference service. \Cref{fig:vAFT_architecture} provides an overview of its architecture.

AGFT is primarily composed of three new core modules: (1) a real-time Contextual Feature Extractor to sense workload changes\cite{Yu2022Challenges}; (2) a decision engine based on a Contextual Bandit, responsible for selecting the optimal frequency\cite{Mishra2020COPE}; and (3) an Adaptive Frequency Controller with intelligent pruning capabilities for safely executing decisions. To achieve its goals, AGFT focuses on three core aspects: real-time workload characterization, reinforcement learning-based frequency decision-making\cite{Mishra2020COPE}, and dynamic and safe action space management.

\subsection{Monitor}

The adaptive decision-making capability of the AGFT framework is built upon a closed-loop perception system that can capture and understand the workload state in real-time. This system achieves precise profiling of the current workload by collecting raw performance metrics from the underlying service and hardware and transforming them into a structured feature vector. This paper uses vLLM\cite{vllm_docs_2025} as a case study to elaborate on the specific design and implementation of this system.\\[3pt]
\textbf{Periodic Metric Acquisition.}
The system's perception layer begins with the Metric Collector module. As illustrated in the \Cref{fig:vAFT_architecture}, this module actively queries vLLM's REST API endpoint \cite{vllm_docs_2025} within a fixed, sub-second sampling period to collect a series of raw performance counters and state values. Although this raw data fully reflects the service's performance, its format and dimensionality are unsuitable for direct decision-making. Therefore, it serves as a time-series data source for the subsequent feature engineering stage.\\[3pt]
\textbf{Workload Feature Engineering and Context Vector Construction.}
The collected raw time-series data is then processed by the Feature Extractor module. The core task of this module is to transform the raw, high-dimensional monitoring data into a low-dimensional, structured representation that accurately reflects the core characteristics of the current workload. This process ultimately generates a 7-dimensional Context Vector, $x_t \in \mathcal{X} = \mathbb{R}^7$. This vector forms the basis of our Contextual Multi-Armed Bandit problem\cite{AGFT_Framework_Proxy,ContextualMAB_Li}, with its dimensions defined as follows:

\begin{enumerate}
    \item Queue Presence: $x_1 = \mathbb{I}[\text{requests\_waiting} > 0]$, a binary indicator for the presence of waiting requests.

    \item Prefill Throughput: $x_2 = \frac{\text{prefill\_tokens}}{\text{sampling\_duration}}$, reflecting the processing speed for new inputs.

    \item Decode Throughput: $x_3 = \frac{\text{decode\_tokens}}{\text{sampling\_duration}}$, reflecting the generation speed for content.

    \item Packing Efficiency: $x_4 = \frac{\text{total\_tokens}}{\text{batch\_iterations}}$, measuring the effectiveness of dynamic batching.

    \item Concurrency: $x_5 = \text{requests\_running}$, representing the number of active concurrent requests.

    \item GPU Cache Usage: $x_6 = \frac{\text{cache\_used\_bytes}}{\text{cache\_total\_bytes}}$, an indicator of GPU memory pressure.

    \item Cache Hit Rate: $x_7 = \frac{\text{cache\_hits}}{\text{cache\_hits}+\text{cache\_misses}}$, measuring the effectiveness of the KV cache mechanism.
\end{enumerate}
Notably, these raw performance metrics are all sourced from vLLM's Prometheus exporter, ensuring our data collection is minimally invasive. We adopt a pure contextual design, which means the context vector $x_t$ does not contain any frequency-related features. In our model, frequency is strictly treated as an Action, rather than as contextual information. This design choice enables the algorithm to learn frequency-agnostic workload patterns while avoiding potential feedback loop issues.

\subsection{Frequency Selection}
The system completes the collection and processing of raw monitoring metrics to generate a 7-dimensional context vector, $x_t$, which characterizes the workload state in real-time\cite{DRLCap_Methodology}. This vector is fed into our core decision-making engine: a Contextual Multi-Armed Bandit (MAB) model customized for GPU frequency optimization\cite{EnergyUCB_MAB}. In this framework, each GPU frequency is an "action," and the system's objective is to learn a policy that selects the action maximizing a reward signal derived from the Energy-Delay Product (EDP) for a given workload "context" $x_t$, while adhering to Service Level Objectives (SLOs). The decision process cyclically executes and intelligently transitions between an Exploration Phase and an Exploitation Phase based on the model's maturity\cite{LinUCB_Foundations}.\\[3pt]
\textbf{Exploration Phase.} During its initial learning phase, the system balances exploration and exploitation using the Upper Confidence Bound (UCB) principle\cite{Ban2024Neural,HCB2024}. The action selection is guided by the LinUCB algorithm\cite{Wang2024Pure,LNUCBTA2025}, which selects the frequency $f_t$ that maximizes the sum of the predicted reward and an exploration bonus based on historical contexts\cite{Kassraie2022,Wang2024Pure}:
\begin{equation}
f_t = \arg \max_{f \in F} \left( \theta_f^T x_t + \alpha_t \sqrt{x_t^T A_f^{-1} x_t} \right)
\end{equation}
This approach prevents premature convergence and allows adaptation to dynamic workload changes.\\[3pt]
\textbf{Exploitation Phase.} Once the model's reward sequence stabilizes, detected via a Page-Hinkley Test (e.g., p-value < 0.05), the system transitions to a pure exploitation phase. In this mature state, it employs a greedy policy, selecting the action with the highest predicted reward for the context $x_t$:
\begin{equation}
f^* = \arg \max_{f \in F_{\text{available}}} \theta_f^T x_t
\end{equation}
This deterministic mode ensures the system consistently applies its learned optimal policy.\\[3pt]
\textbf{Reward Calculation and Model Update.} After executing action $f_t$, a reward $r_t$ is calculated, which is inversely proportional to the measured EDP. This feedback is used to update the model parameters for the executed action according to the LinUCB formulation:
\begin{figure}
    \centering
    \includegraphics[width=1\linewidth]{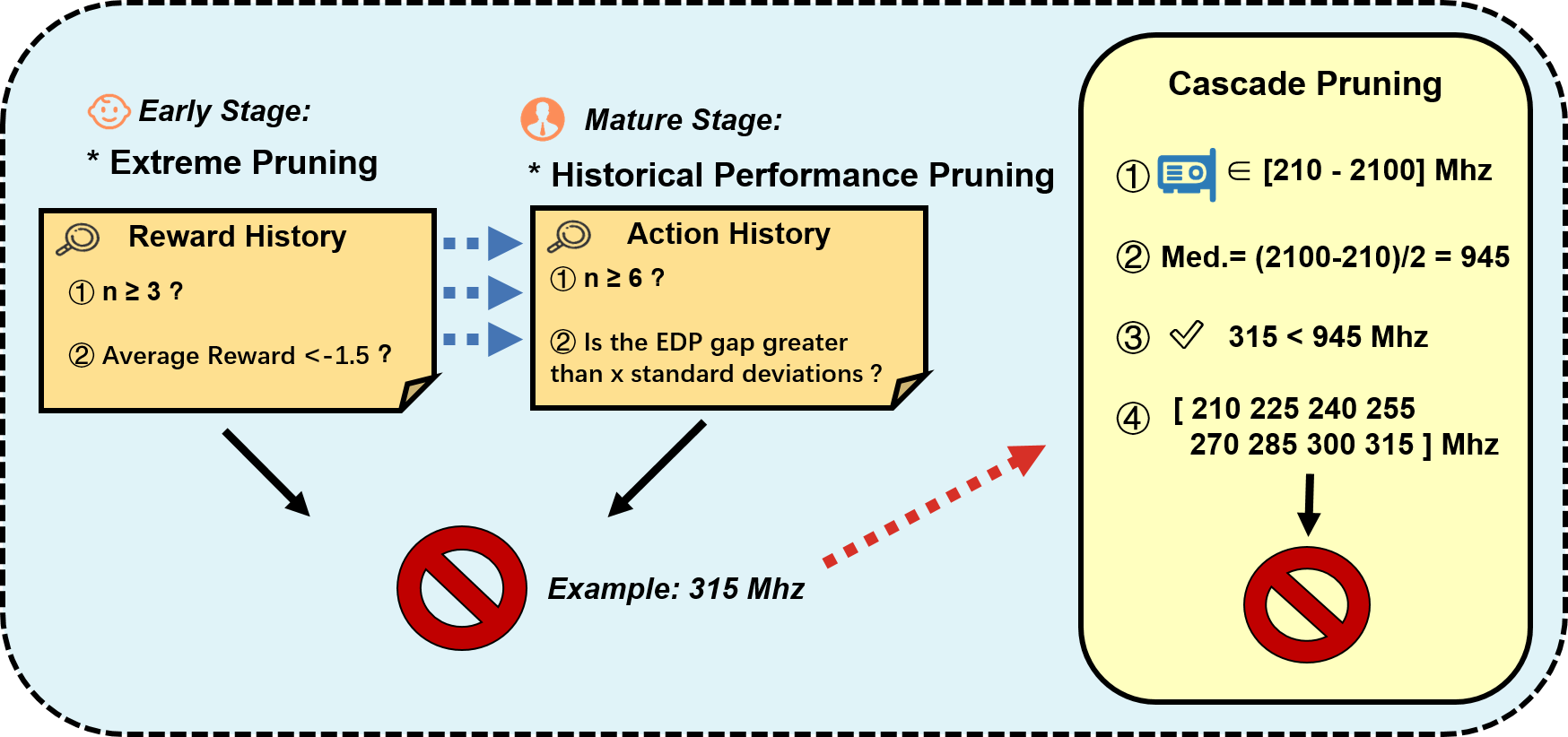}
    \caption{The intelligent pruning framework.}
    \label{fig:9}
\end{figure}
\begin{align}
A_{f_t} &\leftarrow A_{f_t} + x_t x_t^T \\
b_{f_t} &\leftarrow b_{f_t} + r_t x_t
\end{align}
The weight vector $\theta_{f_t}$, representing the learned policy, is then recalculated:
\begin{equation}
\theta_{f_t} = A_{f_t}^{-1} b_{f_t}
\end{equation}
This iterative process enables AGFT to continuously refine its decision-making in dynamic environments.

\subsection{Frequency Pruning}
To enhance learning efficiency and adapt the decision-making process to the characteristics of the underlying hardware, AGFT integrates an Intelligent Pruning Framework. As illustrated in Figure \Cref{fig:9}, this framework dynamically refines the set of available frequency actions (the “action space”) through three complementary mechanisms, ensuring that the learning agent focuses its exploration on more promising regions and thereby accelerates convergence.\\[3pt]
\textbf{Extreme Frequency Instant Pruning. }This mechanism acts as an “early-stage filter” designed for the initial phase of the learning process, aiming to rapidly eliminate frequencies that are unequivocally detrimental. During the initial learning stage (e.g., within the first 60 rounds as per the configuration), when a frequency’s sample count reaches a minimum requirement (e.g., $n_f \ge 3$) and its average reward falls below a strict, negative hard threshold (e.g., $\bar{r}_f < -1.2$), the frequency is identified as a “pathological” case and is immediately and permanently removed from the action space.\\[3pt]
\textbf{Historical Performance Pruning. }After the model passes its initial learning phase and enters a mature stage (e.g., after 30 rounds as configured), this mechanism is activated. A frequency action is only considered for pruning after it has been sufficiently explored (e.g., $n_f \ge 6$) to ensure a statistically reliable performance estimate. Its core logic involves comparing the historical average performance (measured by EDP) of an action against that of the current best-performing action. If its performance is significantly worse, and the gap exceeds a dynamic tolerance threshold calculated from the standard deviation of all actions' performance, the action is deemed suboptimal and pruned.\\[3pt]
\textbf{Cascade Pruning. }The Cascade Pruning mechanism builds upon the two preceding strategies and applies to both, enabling more efficient pruning. It operates on a core heuristic: if a relatively low-frequency action is proven to be suboptimal, it is highly probable that all frequencies below it are also inefficient.

Specifically, when a frequency is pruned (triggered by either Extreme or Historical Pruning) and its value is below a dynamic threshold based on the GPU's hardware capabilities (e.g., half of its maximum supported frequency), the system not only removes that frequency but also cascades the operation, pruning all actions with a lower frequency value from the action space in a single step. This inference mechanism, based on physical intuition, significantly reduces the action space to be explored, focusing the learning process on the most relevant performance range.

\subsection{Mixed Maturity-Based Refinement}
To accelerate the model's convergence process and enhance learning efficiency, the system employs a Mixed Maturity-Based Refinement strategy. The core idea of this strategy is to dynamically adjust its exploration and optimization approach based on the learner's maturity. The system recognizes that the model's predictive capability improves as it interacts more with the environment. Therefore, the system automatically transitions between two distinct refinement modes, Statistical and Predictive, based on whether the number of completed decision rounds has surpassed a predefined learner maturity threshold (e.g., 100 rounds).\\[3pt]
\textbf{Statistical Refinement. }During the initial phase of learning ($t < t_{\text{mature}}$), the model has not yet collected sufficient interaction samples, leading to low confidence in its predictions. To avoid decision-making errors stemming from an immature model, the system employs a more robust Statistical Refinement strategy, which relies entirely on observed historical energy-efficiency data.
\begin{figure}
    \centering
    \includegraphics[width=1\linewidth]{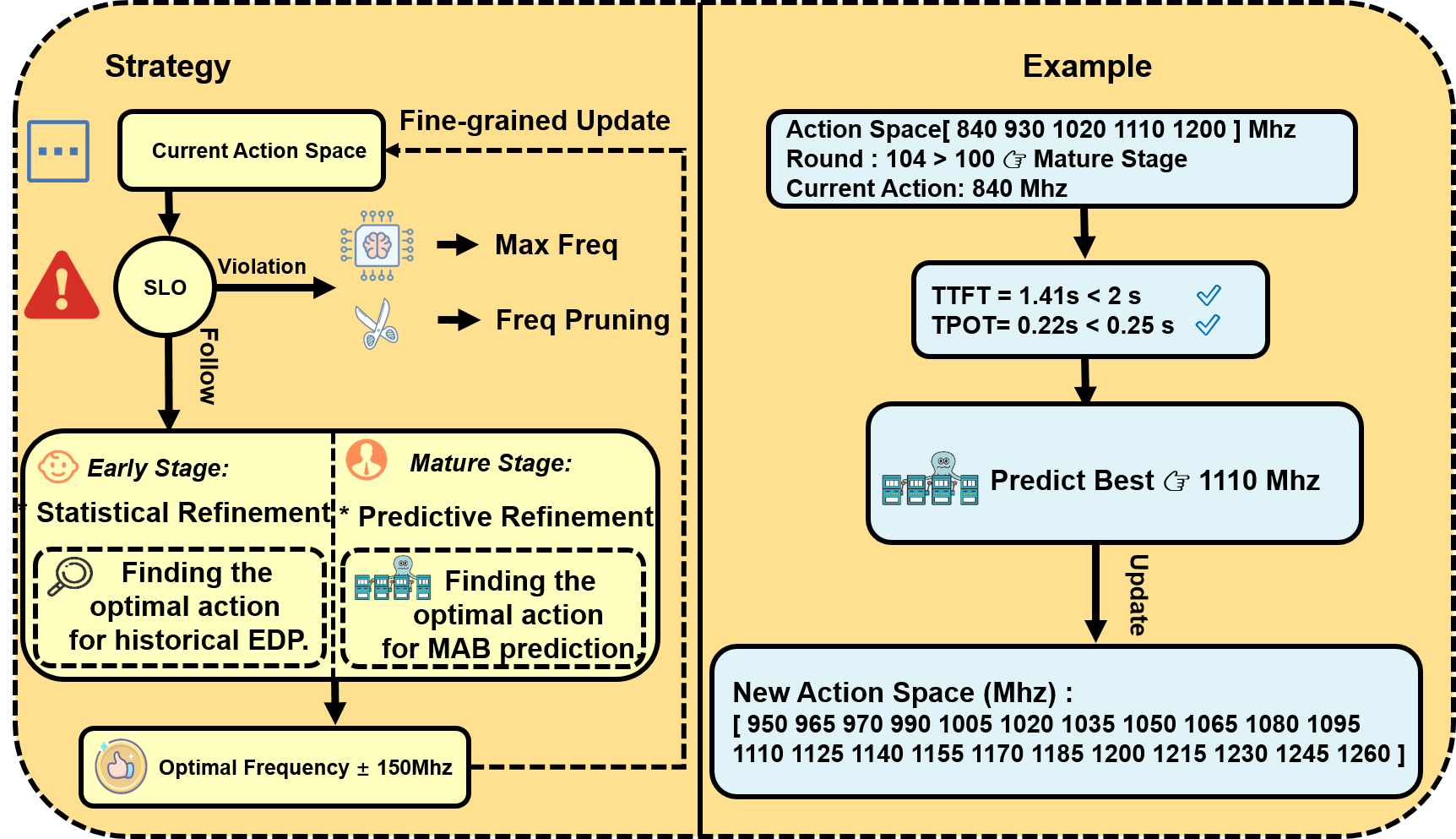}
    \caption{The Mixed Maturity-Based Refinement strategy. }
    \label{fig:adaptive}
\end{figure}
In this phase, the system selects the frequency with the lowest historical average Energy-Delay Product (EDP) as the current "historical optimal anchor" ($f_{\text{anchor}}$), contingent upon meeting a minimum sample requirement (e.g., 4 samples). After determining this historical optimal anchor, the system performs a fine-grained update of the action space, as illustrated in \Cref{fig:adaptive}. Specifically, the system generates a new, high-density action space centered around $f_{\text{anchor}}$ with a range of \textpm 150 MHz and a step size of 15 MHz. This conservative strategy of "empirical validation followed by focused exploration" establishes a reliable performance baseline for the subsequent model-based learning phase.\\[3pt]
\textbf{Predictive Refinement. }Once the learning process surpasses the predefined maturity threshold ($t \ge t_{\text{mature}}$), the system considers the LinUCB model to be sufficiently trained to make reliable predictions based on the current context. At this point, the system transitions to a more proactive and intelligent Predictive Refinement strategy.

In this phase, the core decision-making is entrusted to the mature LinUCB model. The model combines the current real-time context vector, $x_t$, which encapsulates workload features such as queue state and concurrency, and adheres to the principles of the UCB (Upper Confidence Bound) algorithm to evaluate the overall potential of each candidate frequency (i.e., combining expected reward and exploration value).

The system ultimately selects the frequency with the highest overall potential (UCB value) as the optimal anchor ($f_{\text{anchor}}$) for the current refinement. Once the anchor is selected, the system similarly generates a new action space with a range of \textpm 150 MHz and a step size of 15 MHz around that point, thereby focusing exploration resources on the high-reward region predicted by the model.\\[3pt]
Through this transition from passive statistics to model-based prediction, the system can more accurately perceive the characteristics of the current workload and dynamically focus exploration resources on the frequency intervals with the highest expected reward, significantly accelerating convergence to the optimal policy.

\section{Evaluation}
\subsection{Experimental Design }
Our experimental evaluation was conducted on a server equipped with an NVIDIA A6000 GPU. To ensure the relevance and representativeness of our findings, we utilized a workload derived from a 20\% random sampling of the Azure 2024 LLM conversational inference trace. The inference tasks were performed using the Llama-3-3B model. As a performance benchmark, we established a baseline corresponding to the default system configuration, where the GPU operates at its standard, unlocked clock frequencies managed by the native driver. We comprehensively compare our proposed method against this baseline across several key metrics: Service Level Objectives (SLOs), specifically Time-to-First-Token (TTFT) and Time-Per-Output-Token (TPOT); total energy consumption; and the Energy-Delay Product (EDP). Furthermore, to rigorously validate the contributions of our core components, we performed an ablation study by systematically disabling the intelligent action pruning and the adaptive frequency refinement mechanisms.
\begin{figure}
    \centering
    \includegraphics[width=1\linewidth]{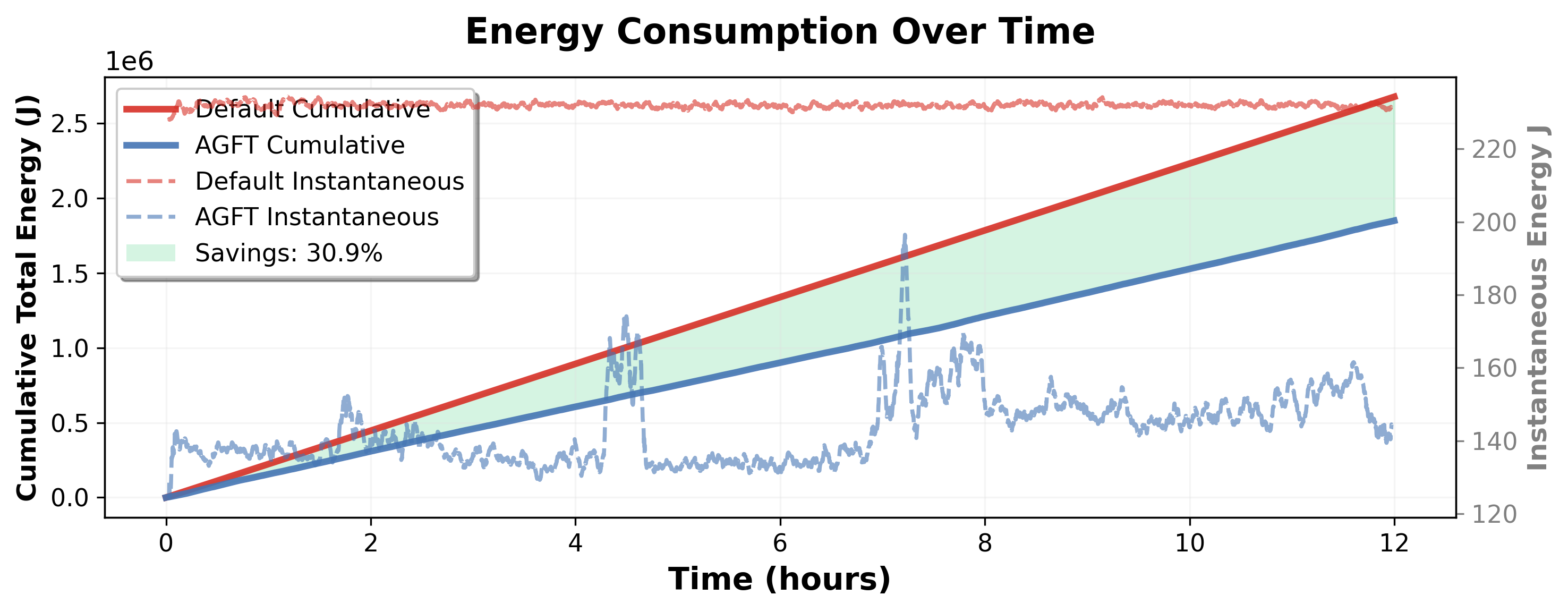}
    \caption{Cumulative energy consumption over 12-hour vLLM inference workload. Solid lines show cumulative values, dashed lines show
  instantaneous rates. AGFT demonstrates significant energy savings compared to static frequency control. }
    \label{fig:11}
\end{figure}
\begin{figure}
    \centering
    \includegraphics[width=1\linewidth]{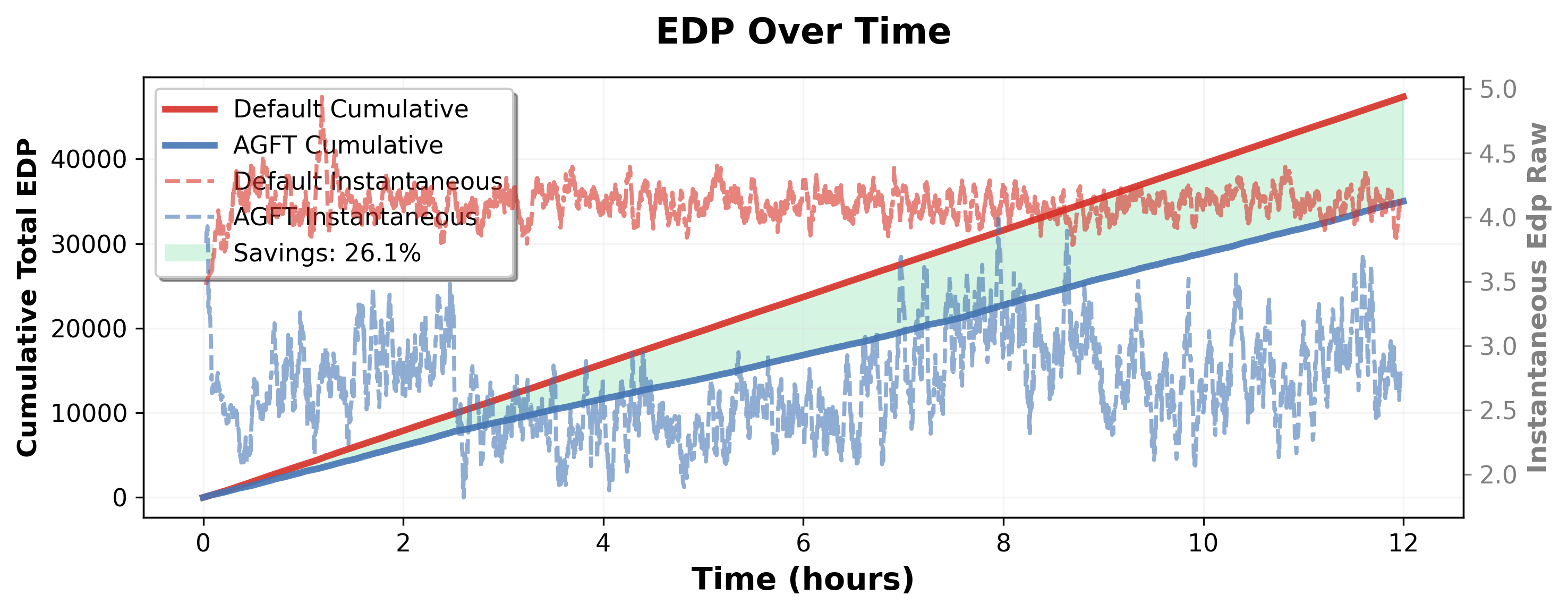}
    \caption{Cumulative Energy-Delay Product (EDP) during 12-hour operation. Solid lines represent cumulative EDP, dashed lines show
  instantaneous values. AGFT substantially reduces total EDP versus static frequency baseline. }
    \label{fig:12}
\end{figure}
\subsection{Long-Term Experiment Overview.}
The results of our extended evaluation, conducted  an 12-hour period driven by a simulated production workload trace, demonstrate the sustained efficiency and effectiveness of our proposed AGFT  framework. The figures illustrate that under these realistic conditions, AGFT consistently outperforms the default baseline in key energy-efficiency metrics.

\Cref{fig:11} and \Cref{fig:12} illustrate the sustained efficiency of the AGFT framework over the 12-hour experiment. \Cref{fig:11} plots the energy consumption, showing AGFT's cumulative usage (blue line) is significantly lower than the default baseline (red line), culminating in a total energy saving of 30.9\%. Similarly, \Cref{fig:12} demonstrates a superior balance between performance and energy, with AGFT achieving a cumulative EDP reduction of 26.1\%. The instantaneous plots in both figures reveal that these savings are primarily attributed to AGFT's dynamic frequency adjustments, which maintain a lower and more adaptive power state compared to the static high-power mode of the baseline.
\begin{table}[t!]
\centering
\caption{Performance Metrics during the Learning Phase (Pre-convergence)}
\label{tab:pre_convergence}
\begin{tabular}{@{}lccc@{}}
\toprule
\textbf{Metric} & \textbf{AGFT mean} & \textbf{Normal mean} & \textbf{Diff} \\ \midrule
Energy (J)       & 130.780                & 230.227              & \textbf{-43.195 \%}        \\
EDP         & 2.889                  & 3.726                & \textbf{-22.443 \%}        \\
TTFT       & 0.048                  & 0.030                & +57.425 \%                 \\
TPOT       & 0.023                  & 0.016                & +40.913 \%                 \\
E2E         & 2.426                  & 1.584                & +53.141 \%                 \\ \bottomrule
\end{tabular}
\end{table}
\begin{figure}[!t]
    \centering 

    \begin{subfigure}{\linewidth}
        \centering
        \includegraphics[width=1\linewidth]{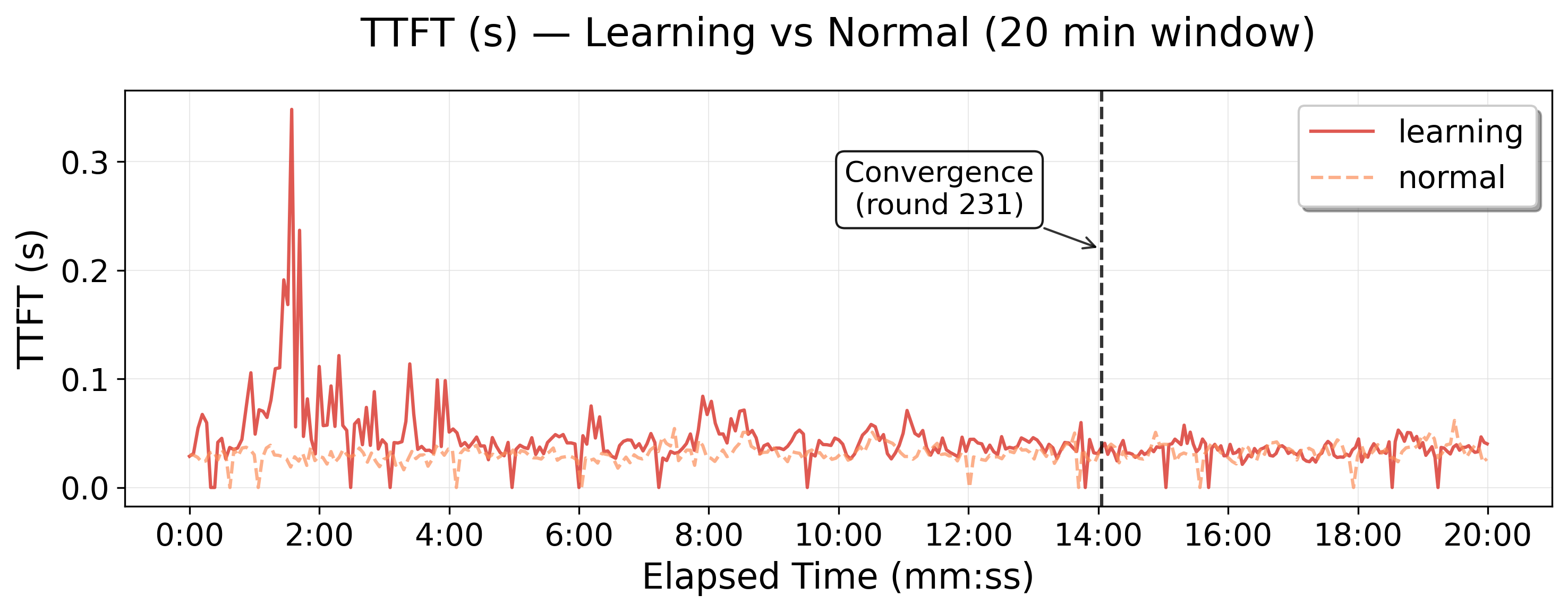}
        \caption{Time-series of Time-to-First-Token (TTFT) in the 20-minute analysis window.}
        \label{fig:ttft_timeseries}
    \end{subfigure}
    
    \begin{subfigure}{\linewidth}
        \centering
        \includegraphics[width=1\linewidth]{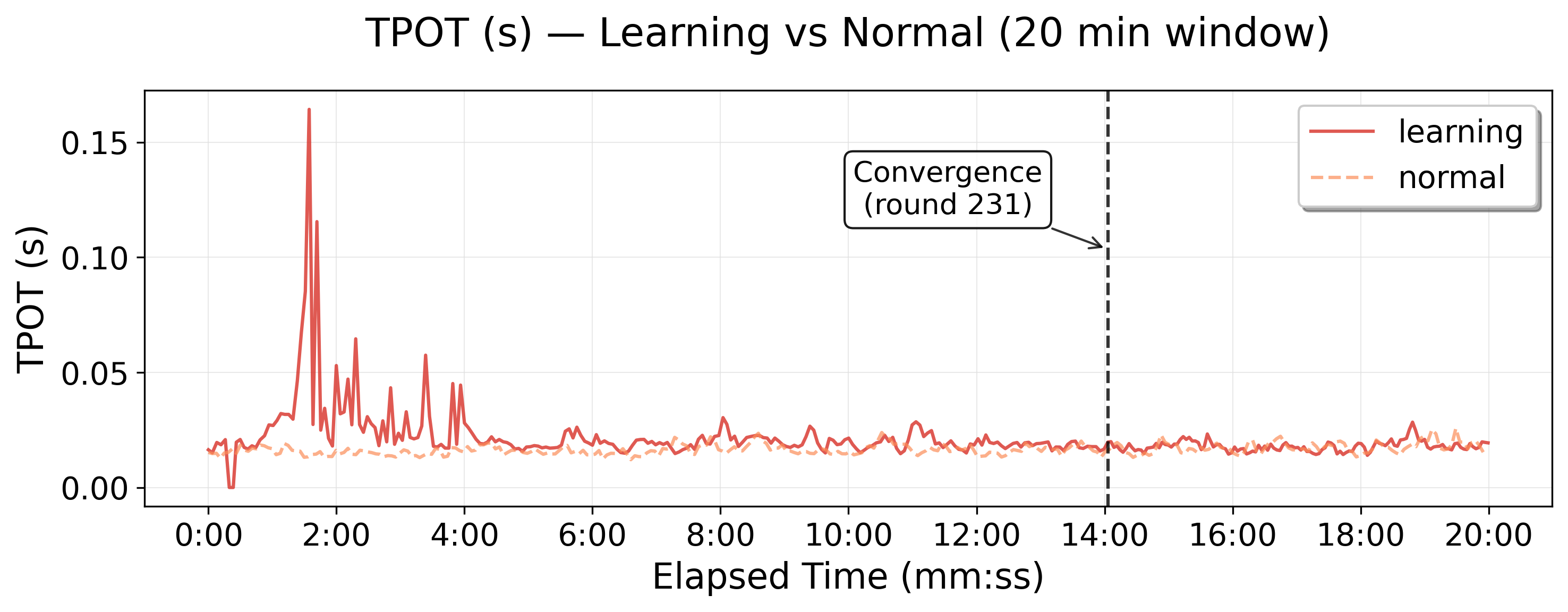}
        \caption{Time-series of Time-Per-Output-Token (TPOT) in the 20-minute analysis window.}
        \label{fig:tpot_timeseries}
    \end{subfigure}
    
    \begin{subfigure}{\linewidth}
        \centering
        \includegraphics[width=1\linewidth]{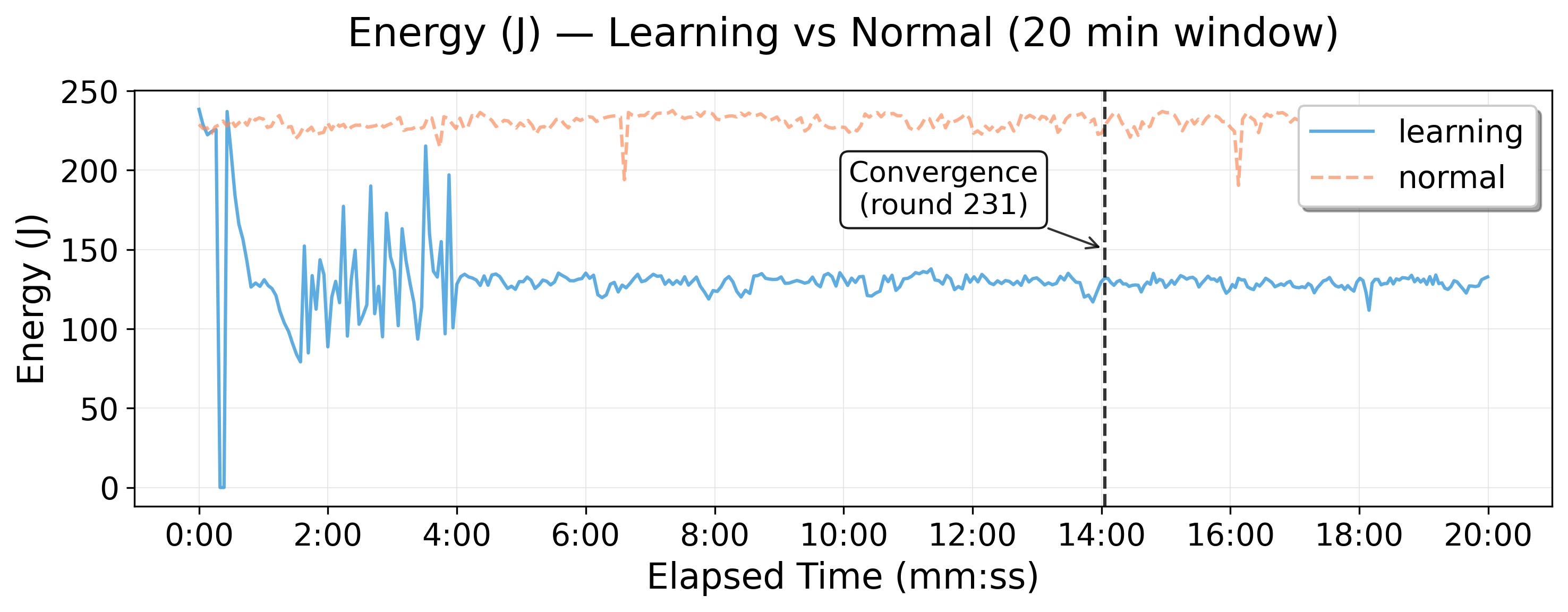}
        \caption{Time-series of Energy Consumption (J) in the 20-minute analysis window.}
        \label{fig:energy_timeseries} 
    \end{subfigure}
    
    \begin{subfigure}{\linewidth}
        \centering
        \includegraphics[width=1\linewidth]{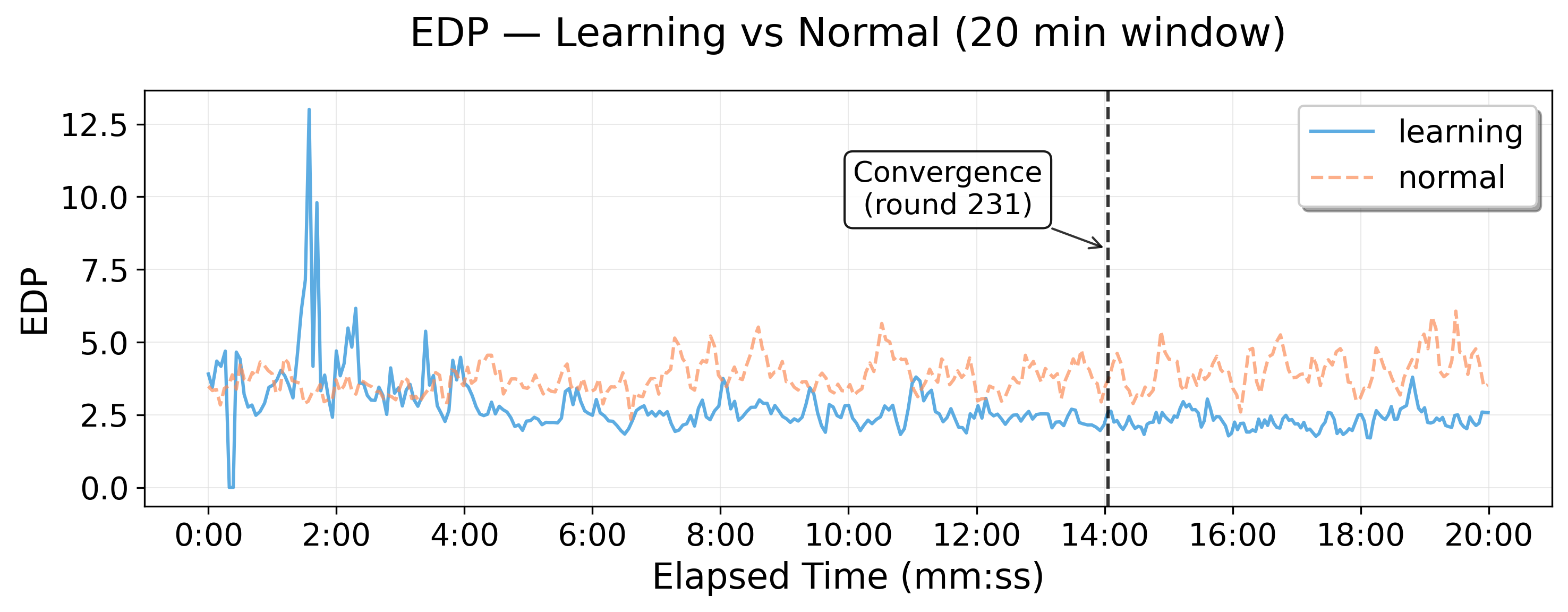}
        \caption{Time-series of Energy-Delay Product (EDP) in the 20-minute analysis window.}
        \label{fig:edp_timeseries} 
    \end{subfigure}

    \caption{Time-series comparison of key performance metrics in the 20-minute analysis window.}
    \label{fig:performance_comparison} 
\end{figure}
Furthermore, \Cref{fig:12} highlights the comparison of the Energy-Delay Product (EDP), a critical metric for overall system efficiency. Our AGFT approach yields an average EDP reduction of 34.6\%, which underscores its ability to achieve a superior balance between performance and energy consumption throughout the long-duration test.
\label{sec:implementation}

\subsection{Detailed Analysis}
For a fine-grained analysis of the system's behavior, we examine the initial 20-minute operational window, which captures the AGFT framework's transition from an initial learning phase to a stable, post-convergence phase. This analysis reveals the nuanced trade-offs our system makes to achieve its final, efficient state. The performance metrics for the pre- and post-convergence periods are summarized in \Cref{tab:pre_convergence,tab:post_convergence}.\\[3pt]\begin{figure}
    \centering
    \includegraphics[width=1\linewidth]{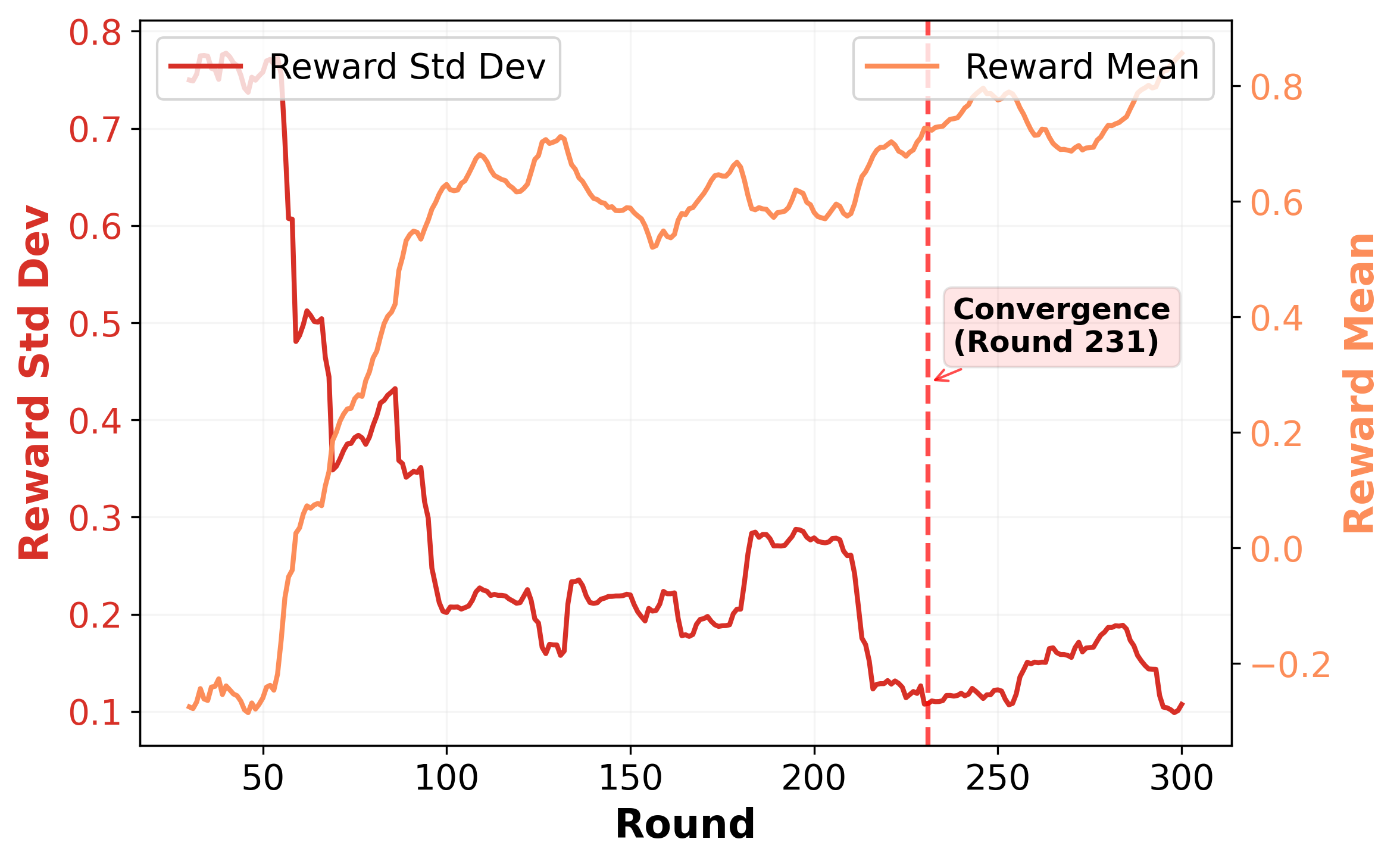}
    \caption{Reward statistics evolution during contextual multi-armed bandit learning. Rolling standard deviation (red) and mean (orange) of rewards 
  demonstrate the algorithm's progression from exploration to stable exploitation of optimal GPU frequency policies.}
\label{fig:convergence_indicators}
\end{figure}\
\textbf{Learning Phase Trade-offs}. During the initial learning phase (approximately the first 14 minutes), the system prioritizes exploration to identify optimal operational policies. This exploratory behavior intentionally trades higher latency for greater insight into the system's dynamics. As shown in \Cref{tab:pre_convergence}, this results in a noticeable increase in average TTFT by +57.4\% and TPOT by +40.9\% compared to the stable baseline. The volatility associated with this phase is also clearly visible in the time-series plots for TTFT (\Cref{fig:ttft_timeseries}) and TPOT (\Cref{fig:tpot_timeseries}). However, this short-term latency cost is immediately compensated by remarkable energy savings, with average energy consumption reduced by a substantial 43.2\% (\Cref{fig:energy_timeseries}). This ultimately leads to an immediate and significant improvement in the Energy-Delay Product (EDP) by 22.4\%, as illustrated in \Cref{fig:edp_timeseries}, even during this volatile learning period.\\[3pt]
\textbf{Post-convergence Efficiency}. Following convergence at round 231, the system transitions into an exploitation phase, where it leverages the learned policy to maximize efficiency. The data in \Cref{tab:post_convergence} demonstrates the success of this learned policy. The latency overhead is dramatically reduced to a minimal +9.3\% for TTFT and +7.1\% for TPOT, while throughput remains nearly identical to the baseline (-2.2\% difference). Crucially, the substantial energy savings are maintained, with a reduction of 44.3\%. As shown in \Cref{fig:edp_timeseries}, this combination of minimal latency impact and sustained energy reduction results in a much more pronounced and stable improvement in EDP, which is now 40.3\% lower than the baseline. This clearly validates our framework's ability to not only learn but also apply a highly efficient and stable operational strategy in the long term.\\[3pt]\begin{table}[t!]
\centering
\caption{Performance Metrics in the Stable Phase (Post-convergence)}
\label{tab:post_convergence}\begin{tabular}{@{}lccc@{}}
\toprule
\textbf{Metric}          & \textbf{AGFT mean} & \textbf{Normal mean} & \textbf{Diff} \\ \midrule
Energy (J)                & 129.058                & 231.626              & \textbf{-44.282 \%}        \\
EDP                  & 2.427                  & 4.065                & \textbf{-40.287 \%}        \\
TTFT                & 0.037                  & 0.033                & +9.271 \%                  \\
TPOT                & 0.019                  & 0.018                & +7.118 \%                  \\
E2E                  & 2.096                  & 1.960                & +6.922 \%                  \\
\bottomrule
\end{tabular}
\end{table}
\begin{table*}[t] 
\centering
\caption{Ablation Study: Disabling Fine-Grained Frequency Control ("No-grain")}
\label{tab:ablation_nograin}
\begin{tabular*}{\textwidth}{@{\extracolsep{\fill}}l r r r r r r@{}} 
\toprule
\textbf{Metric} & \textbf{Normal} & \textbf{No-grain} & \textbf{Diff} & \textbf{CV} & \textbf{CV} & \textbf{\ CV} \\ 
& \textbf{Mean} & \textbf{Mean} & & \textbf{Normal} & \textbf{No-grain} & \textbf{Diff} \\
\midrule
Energy (J)       & 129.06   & 130.70   & \textbf{+1.27\,\%}   & 0.029  & 0.073  & \textbf{+151\,\%} \\
EDP         & 2.43     & 2.65     & \textbf{+9.24\,\%}   & 0.178  & 0.238  & +34\,\%           \\
TTFT       & 0.0365   & 0.0395   & +8.02\,\%            & 0.292  & 0.409  & +40\,\%           \\
TPOT       & 0.0188   & 0.0202   & +7.85\,\%            & 0.158  & 0.225  & +43\,\%           \\
E2E         & 2.10     & 2.30     & +9.53\,\%            & 0.976  & 0.992  & +1.6\,\%          \\ \bottomrule
\end{tabular*}
\end{table*}
\begin{table}[h!] 
\centering
\caption{Ablation Study: Disabling Action Space Pruning ("No pruning")}
\label{tab:ablation_nopruning}
\begin{tabular}{@{}l r r r@{}}
\toprule
\textbf{Metric}               & \textbf{CV Normal} & \textbf{CV No pruning} & \textbf{Diff} \\ \midrule
Energy (J)                     & 0.208      & 0.235          & -11.4\,\%        \\
EDP                       & 0.409      & 0.612          & \textbf{-33.1\,\%} \\
TTFT                     & 0.696      & 0.762          & -8.7\,\%         \\
TPOT                     & 0.631      & 0.921          & \textbf{-31.5\,\%} \\
E2E                       & 1.086      & 1.054          & +3.1\,\%         \\
         \bottomrule
\end{tabular}
\end{table}
\textbf{Online Learning Efficacy.} The efficacy of the online learning process is validated by analyzing the agent's reward statistics evolution, as presented in \Cref{fig:convergence_indicators}. The plot illustrates the agent's progression from an initial exploration phase to a stable exploitation phase. In the early stage (approximately before round 100), the high standard deviation of the reward (red line) reflects significant uncertainty and exploratory behavior, while the rolling mean reward (orange line) is low. As the agent accumulates experience, a clear learning trend emerges: the standard deviation consistently decreases while the mean reward steadily climbs, indicating the agent is successfully identifying and favoring better actions. After the convergence point (round 231), both metrics stabilize, with the mean reward remaining high and its standard deviation low. This transition from high uncertainty to high certainty strongly validates that our online learning approach successfully converges to a stable and effective policy.\\[3pt]
\begin{table}[t!]
\centering
\small
\caption{Comparison of theoretically optimal frequencies (Offline) versus those learned by AGFT (Online) across workload prototypes.}
\label{tab:freq_comparison}
\begin{tabular}{l S[table-format=4.0] S[table-format=4.0] S[table-format=+1.1]}
\toprule
\textbf{Workload} & {\textbf{Offline}} & {\textbf{Online}} & {\textbf{Deviation}} \\
\midrule
Normal           & 1230 & 1230 & +0.0\% \\
Long Context     & 1395 & 1410 & +1.1\% \\
Long Generation  & 1260 & 1200 & -4.8\% \\
High Concurrency & 1365 & 1320 & -3.3\% \\
High Cache Hit   & 1200 & 1290 & +7.\%5 \\
\bottomrule
\end{tabular}
\end{table}
\textbf{Comparative Analysis against the Theoretical Optimum.} The data presented in Table \ref{tab:freq_comparison} clearly demonstrates the efficacy of our online learning method. This was evaluated across the five distinct workloads detailed in Table \ref{tab:workload_config}. For each scenario, the system processed 5,000 requests to ensure our learning algorithm had sufficient opportunity to converge. The learned frequencies align closely with the theoretical optimal values derived from offline benchmarking. Specifically, under the Normal workload, the algorithm precisely identifies the optimal frequency of 1230 MHz (0\% deviation), showcasing perfect adaptability in a baseline scenario. For more demanding, compute-intensive workloads such as Long Context and High Concurrency, the algorithm selects frequencies with minimal deviation from the optimal values (+1.1\% and -3.3\%, respectively), proving its capability to effectively perceive and adapt to high-demand dynamic conditions. Even with slightly larger deviations under the Long Generation (-4.8\%) and High Cache Hit (+7.5\%) workloads, the results remain well within an acceptable range, which thoroughly demonstrates the robustness and effectiveness of our proposed method across varied and complex operational states.

\subsection{Ablation Study}
To quantitatively assess the individual contributions of our framework's core components, we conduct an ablation study. We analyze two specific scenarios: one where the fine-grained frequency control is disabled, and another where the intelligent action space pruning is turned off. Both experiments are compared against our full AGFT framework over a 20-minute operational window.\\[3pt]
\textbf{Impact of Fine-Grained Frequency Control. } In this experiment, we disable the ability of our agent to select fine-grained frequency steps, forcing it to use a coarser action space ("No-grain"). The results, summarized in \Cref{tab:ablation_nograin}, demonstrate that fine-grained control is critical for both optimality and stability. The mean performance metrics show a notable degradation without it, with average EDP increasing by +9.24\% and energy consumption by +1.27\%. The impact on system stability is even more pronounced. The Coefficient of Variation (CV), which measures relative volatility, increases dramatically for energy (+151\%) and EDP (+34\%). This indicates that the fine-grained action space is essential for the agent to make precise, stable adjustments, preventing oscillatory and inefficient behavior.\\[3pt]
\textbf{Impact of Action Space Pruning. } Next, we evaluate the effectiveness of our intelligent pruning mechanism by disabling it entirely ("No pruning"). The results in \Cref{tab:ablation_nopruning} focus on the impact on system stability, measured by the Coefficient of Variation. The data shows that removing the pruning mechanism leads to a significant increase in the volatility of key metrics. For instance, the CV for EDP and TPOT is substantially higher in the "No pruning" configuration. This highlights the crucial role of our pruning strategy in stabilizing the learning process by eliminating suboptimal actions early, thereby allowing the agent to converge on a more consistent and reliable policy.

\section{Conclusion}
\label{sec:conclusion}

We introduced AGFT, an adaptive GPU frequency optimization framework for modern LLM inference services. It pioneers a closed-loop, online learning approach that eliminates the reliance on offline modeling. At its core, AGFT employs a minimally intrusive workload characterization method, using a 7-dimensional  'fingerprint' to identify distinct workload prototypes in real-time. This privacy-preserving technique allows the system to adapt to workload fluctuations in dynamic scheduling environments, such as those using continuous batching, without accessing user prompt content. By intelligently optimizing the EDP while guaranteeing SLOs, AGFT demonstrates a robust path towards more energy-efficient and privacy-aware AI infrastructure.
\begin{acks}
We thank the vLLM development team for their excellent inference framework and the anonymous reviewers for their valuable feedback. We acknowledge the computational resources provided by our institution and the contributions of the open-source community in developing the foundational tools that enabled this research. Special thanks to the NVIDIA developer community for GPU control APIs and documentation.
\end{acks}

\bibliographystyle{ACM-Reference-Format}
\bibliography{references}
\end{document}